\documentclass[conference]{IEEEtran}

\usepackage{siunitx}
\usepackage{hyperref}
\usepackage{mathtools}
\usepackage{epsfig, amsmath}
\usepackage{xcolor}
\usepackage{algorithm}
\usepackage{algpseudocode}
\usepackage{multirow}
\usepackage{subcaption}
\usepackage{comment}
\usepackage{tikzpagenodes}

\usepackage{microtype}

\usepackage{mdframed} 
\newmdtheoremenv[
linecolor=gray,leftmargin=0,
rightmargin=0,
backgroundcolor=gray!10,
ntheorem]{myprop}{Take-away}[section]

\usepackage[textsize=scriptsize,textwidth=2.8cm,disable]{todonotes}

\begin{document}

\title{Deep k-Nearest Neighbors: Towards Confident, Interpretable and Robust Deep Learning }
\author{
	Nicolas Papernot and Patrick McDaniel\\
	Department of Computer Science and Engineering, Pennsylvania State University\\
	\{ngp5056, mcdaniel\}@cse.psu.edu}

\maketitle

\thispagestyle{plain}
\pagestyle{plain}

\IEEEpeerreviewmaketitle

\begin{abstract}
Deep neural networks (DNNs) enable innovative applications of machine learning
like image recognition, machine translation, or malware detection. 
However, deep learning
is often criticized for its lack of robustness in adversarial settings (e.g., vulnerability to adversarial inputs) and general inability to rationalize its predictions. 
In this work, we exploit the structure of deep
learning to enable new learning-based inference and decision strategies that achieve desirable properties such as robustness and interpretability. 
We take a first step in this 
direction and 
introduce the Deep k-Nearest Neighbors (DkNN).
This hybrid classifier combines the k-nearest neighbors algorithm with representations of the data learned by each layer of the DNN: a test input is compared to its
neighboring training points according to the distance that
separates them in the representations.
We show 
the labels of these neighboring points
afford confidence estimates for inputs outside
the model's training manifold, including on malicious inputs like adversarial examples--and therein provides protections against inputs that are outside the models understanding. 
This is because the nearest neighbors can be used to estimate
the nonconformity of, i.e., the lack of
support for, a prediction in the training data.
The neighbors also constitute human-interpretable explanations of predictions.
We evaluate the DkNN algorithm on several datasets, and show the confidence estimates accurately identify inputs outside the model, and that the explanations provided by nearest neighbors are intuitive and useful in understanding model failures.
\end{abstract} \section{Introduction}
\label{sec:introduction}

Deep learning is ubiquitous: deep neural networks achieve exceptional
performance on challenging tasks like  machine translation~\cite{sutskever2014sequence, bahdanau2014neural},
diagnosing medical conditions such as diabetic retinopathy~\cite{gardner1996automatic,gulshan2016development}
or pneumonia~\cite{caruana2015intelligible}, malware detection~\cite{grosse2016adversarial, saxe2015deep, zhu2016featuresmith},
and classification of images~\cite{krizhevsky2012imagenet,he2016deep}. This success is often attributed
in part to developments in hardware (e.g., GPUs~\cite{owens2008gpu} and TPUs~\cite{jouppi2017datacenter}) and availability of large datasets (e.g., ImageNet~\cite{deng2009imagenet}),
but more importantly also to the architectural design of neural networks 
and the remarkable performance of stochastic gradient descent. 
Indeed, deep neural networks are designed to learn a hierarchical set
of \textit{representations} of the input domain~\cite{hinton2007learning}. These representations project the input data in increasingly abstract spaces---or
 \textit{embeddings}---eventually sufficiently abstract for the 
task to be solved (e.g., classification) with a linear decision function.

Despite the breakthroughs they have enabled, the adoption of
deep neural networks (DNNs) in security and safety critical
applications remains limited in part because they are often considered
as \textit{black-box} models whose performance is not entirely understood 
and are controlled by a large set of parameters--modern DNN architectures 
are often parameterized with over a million values. This is paradoxical because of the
very nature of deep learning: an essential part of the design philosophy of DNNs
is to learn a modular model whose components (layers of 
neurons) are simple in isolation
yet powerful and expressive in combination---thanks to their orchestration
as a composition of non-linear functions~\cite{hinton2007learning}. 

In this paper, we harness this intrinsic
modularity of deep learning to address three well-identified criticisms directly relevant to its security:
the lack of reliable \textit{confidence} estimates~\cite{guo2017calibration},
model \textit{interpretability}~\cite{lipton2016mythos} and 
 \textit{robustness}~\cite{szegedy2013intriguing}. 
We introduce the 
\textit{Deep k-Nearest Neighbors} (DkNN) classification algorithm, which enforces conformity
of the predictions made by a DNN on test inputs  with respect to the model's training data.
For each layer in the DNN, the DkNN performs a nearest neighbor
search to find training points for which the layer's output
is closest to the layer's output on the test
input of interest.
We then analyze the label of these neighboring training points to ensure
that the intermediate computations performed by each layer
remain conformal with the final model's prediction.

In adversarial settings, this yields an approach to defense that differs
from prior work in that it addresses the underlying cause of 
poor model performance on malicious inputs rather 
than attempting to make particular adversarial strategies fail.
Rather than nurturing model integrity by attempting to correctly
classify all legitimate and malicious inputs, we ensure the integrity of the model by creating a novel characterization of confidence, called \textit{credibility}, that spans the hierarchy of representations within of a DNN:
any credible classification
must be supported by evidence from the training data. Conversely, 
a lack of credibility indicates that the sample must be ambiguous or adversarial. Indeed, 
the large error space of ML models~\cite{tramer2017ensemble} exposes a
large attack surface, which is exploited by attacks through threat vectors like
adversarial examples (see below). 

Our evaluation shows that the integrity of the DkNN classifier is maintained 
when its prediction is  supported by the underlying training manifold. This support is evaluated as level of ``confidence'' in the prediction's agreement with the  
nearest neighbors found at each layer of the model and analyzed with conformal prediction~\cite{saunders1999transduction,vovk1999machine}.
Returning to desired properties of the model; (a) confidence can be viewed as estimating the distance between the test input and the model's training points, (b) interpretability is achieved by finding points on the training manifold supporting the prediction, and (c) robustness is achieved when the prediction's support is consistent across the layers of the DNN, i.e., prediction has high confidence. 

\begin{figure*}[t] 
	\centering
	\includegraphics[width=\textwidth]{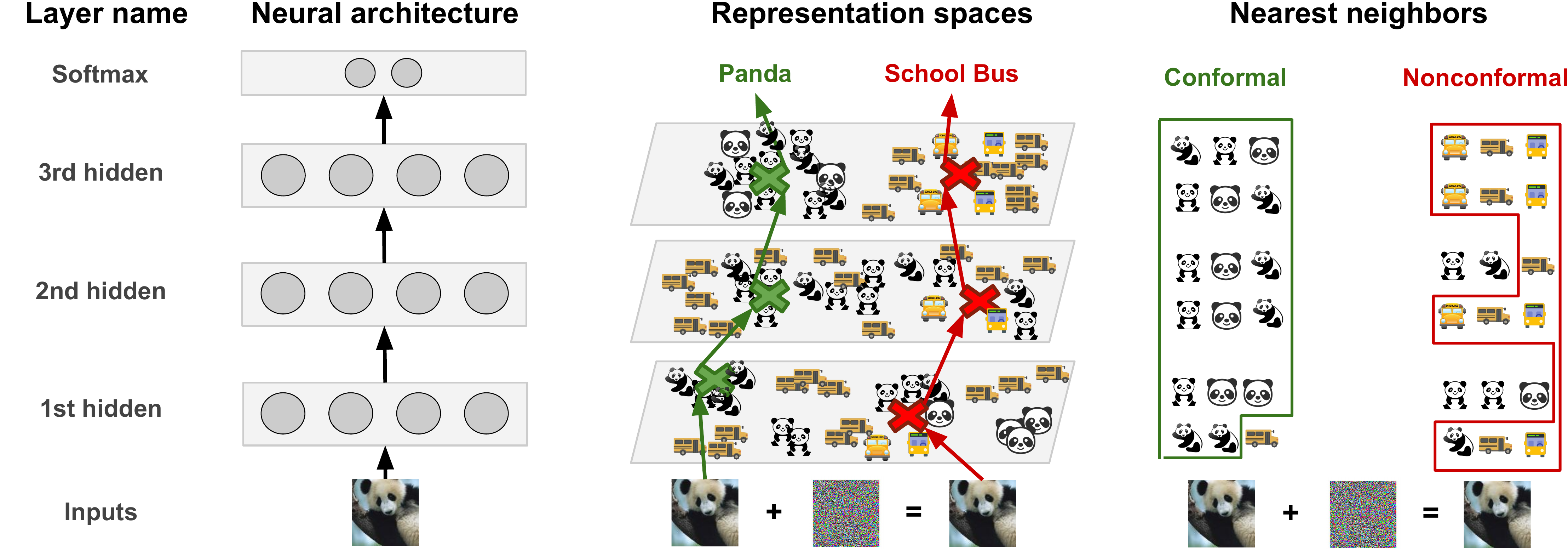}
	\caption{\textbf{Intuition behind the Deep k-Nearest Neighbors (DkNN)}---Consider a  deep neural network (left), representations output by each layer (middle) and the nearest neighbors found at each layer in the training data (right). Drawings of pandas and school buses indicate training points. Confidence is high when there is homogeneity among the nearest neighbors labels (e.g., here for the unmodified panda image). Interpretability of the outcome of each layer is provided by the nearest neighbors. Robustness stems from detecting nonconformal predictions from nearest neighbor labels found for out-of-distribution inputs (e.g., an adversarial panda) across different layers. Representation spaces are high-dimensional but depicted in 2D for clarity. }
	\label{fig:dknn-motivation}
\end{figure*}

\subsection*{Intuition for the Deep k-Nearest Neighbors}
The intuition behind DkNN is presented in Figure~\ref{fig:dknn-motivation}.  Discussed below, this gives rise to explorations of the definition and importance of  confidence, interpretability and robustness and their role in machine learning in adversarial settings.

\paragraph{Confidence}
There have been recent calls from the security and ML communities to more precisely 
calibrate the confidence of 
predictions made by DNNs~\cite{guo2017calibration}. 
This is critical in tasks like  
pedestrian detection for self-driving cars~\cite{bojarski2016end} or automated diagnosis
of medical conditions~\cite{jiang2011calibrating}.
Probabilities output by DNNs are commonly used as a proxy for their
confidence. Yet, these probabilities are not
faithful indicators of model confidence (see Section~\ref{ssec:background-confidence}). 
A notable counter-example is the one of
adversarial examples, which are often classified with more ``confidence'' (per the DNN's output probabilities) than
their legitimate counterpart, despite the model prediction being erroneous on these inputs~\cite{szegedy2013intriguing,biggio2013evasion}.
Furthermore, when a DNN assigns  equal 
probabilities to two candidate labels (i.e., it has low confidence in either outcomes), it may do so for at least two different reasons: (a) the DNN 
has not analyzed similar inputs during training and is 
extrapolating,
or (b) the input is ambiguous---perhaps as a result of an adversary attempting to 
subvert the  system or the sample being collected with a naturally noisy observation process. 

In the DkNN,
 the number of nearest
neighboring training points whose label does not match the prediction
 made on a test input defines an estimate of
 the input's nonconformity to the training data.
 The larger that number is, the weaker the training data supports the prediction.
To formalize this, we operate in the framework of conformal prediction~\cite{shafer2008tutorial}
and compute both the \textit{confidence} and \textit{credibility} of DkNN predictions.
The former quantifies the likelihood of the prediction being correct given 
the training set, while the later characterizes how relevant the training set is
to the prediction. In our experiments (see Sections~\ref{sec:eval-confidence} and~\ref{sec:eval-robustness}), we find that credibility is able to 
reliably identify the lack of support from the training data when predicting
far from the training manifold.

\paragraph{Interpretability}
This property is the ability to construct
an explanation for model predictions that can be
easily understood by a human observer~\cite{doshi2017towards}, or put another way
to rationalize DNN predictions based on evidence---and answer the question:
``Why did the model decide that?''.
DNN decisions are
difficult to \textit{interpret} because neurons 
are arranged in a complex sequence of computations and
the output representation of each layer is high-dimensional.
This
limited interpretability inhibits applications of deep learning
in domains like healthcare~\cite{caruana2015intelligible}, where trust in model predictions is key. 
In contrast, the DkNN algorithm is more interpretable
by design because the nearest neighbors themselves provide explanations,
for individual layer and overall DNN predictions, that are easily understood by humans because they lie in the input domain.

\paragraph{Robustness}
\textit{Robustness} to input perturbation is another important
requirement for security~\cite{dalvi2004adversarial,lowd2005adversarial,barreno2006can,barreno2010security,papernot2016towards}
and safety~\cite{amodei2016concrete} in ML systems. While DNNs are robust
to random perturbations of their inputs, they are vulnerable to 
small intentional perturbations of their inputs at test
time---known as 
\textit{adversarial examples}~\cite{szegedy2013intriguing,biggio2013evasion}.
This attack vector allows an adversary to fully control
a DNN's predictions, even if it has no access to the
model's training data or internal parameters~\cite{papernot2017practical}.
Small perturbations introduced by adversarial examples
are able to arbitrarily change the
DNN's output because they are gradually magnified by non-linearities successively applied by each layer in the model.
Put another way, when the DNN is misclassifying an input, 
there is necessarily one of its layers that transformed the input's
representation, which was by definition in the correct class initially.
In contrast, the DkNN classifier prevents this by identifying changes in the labels of
nearest neighboring training points between lower and higher
layers of the DNN as an indicator that the DNN is mispredicting
(see Figure~\ref{fig:dknn-motivation}).
In essence, the DkNN removes exploitable degrees of freedom
available to adversaries attempting to manipulate the system's
predictions---thus offering a form of robustness to adversarial example attacks (see Section~\ref{sec:eval-robustness}).
Note that this is \textit{not} simply an ensemble approach that
combines the predictions from multiple models; our DkNN algorithm
inspects the intermediate computations of a \textit{single} 
DNN to ensure its predictions are conformal with its training data. 

\subsection*{Contributions}

\noindent To summarize, we make the following contributions:
\begin{itemize}

	\item We introduce the Deep k-Nearest Neighbors (DkNN) algorithm that measures the nonconformity of a prediction on a test input with
	the training data as an indirect estimate of the credibility of the model predictions (see Section~\ref{sec:approach}).
	
	\item We empirically validate that predictions made by a DkNN are more reliable estimates of credibility than a DNN on naturally occurring out-of-distribution inputs. On inputs geometrically transformed or from classes not included in
	the training data, the DkNN's credibility measurement is below $10\%$ versus $20\%$-$50\%$ for a given DNN. 
	\item We demonstrate DkNN interpretability through a study of  racial bias and fairness in a well known DNN (Section~\ref{sec:eval-interpretability}).

	\item We show that the DkNN is able to identify adversarial examples generated using existing algorithms because of their low credibility (see Section~\ref{sec:eval-robustness}). We also show that adaptive attacks against the DkNN often need to perturb input semantics to change the DkNN's prediction.
	
\end{itemize}

We find these results encouraging and note that they highlight the benefit of analyzing confidence, interpretability and robustness as related properties of a DNN.   Here, we exploit the DNN's modularity  and verify the conformity of predictions with respect to training data at each layer of abstraction, and therein ensure that the DNN converges toward a rational and interpretable output.
Interestingly and as explored in Section~\ref{sec:eval-robustness},
Sabour et al.~\cite{sabour2015adversarial} investigated the vulnerability of internal representations as a vehicle for creating malicious inputs.
This suggests that in addition to enforcing these properties at the level of the model as a whole, it is important to defend each abstraction from malicious manipulation. Indeed the work discussed throughout suggests that is is not only necessary, but also a useful tool in providing a potential defense against existing adversarial algorithms. 
  \section{Background on Deep Learning}
\label{sec:background}

\textbf{Machine learning} refers to a set of techniques that automate the analysis
of large scale data. In this paper, we consider \textit{classification} tasks where ML models are
designed to learn mappings between an input domain and a predefined set of
outputs called~\emph{classes}. For instance, the input domain may be PDF files and the 
classes ``benign'' or ``malicious'' when the task of interest is malware detection in
PDF documents. Techniques like \textit{support vector machines}~\cite{cortes1995support},
and more recently \textit{deep learning}~\cite{Goodfellow-et-al-2016}---revisiting
\textit{neural networks} architectures~\cite{hopfield1987neural}
are common choices to learn supervised models from data.

In this paper we build on \textbf{deep neural networks}~\cite{krizhevsky2012imagenet} (DNNs).
DNNs are designed to learn hierarchical---and
increasingly abstract---representations
of the data. For instance, a neural network trained to recognize objects when presented with samples will typically first learn representations
of the images that indicate the presence of various geometric shapes and colors, compose
these to identify subsets of objects before it reaches its final representation which is the prediction~\cite{hinton2007learning}. 
Specifically, a deep neural network $f$ is a composition of $l$ parametric
functions referred to as \textit{layers}. 
Each layer can be seen as a representation of the input domain.
A layer is made up of neurons---small units that compute
 one dimension of the layer's output.
The layer indexed $\lambda$ (with $\lambda\in 0\text{ .. }l-1$)
takes as its input the output of previous layer $f_{\lambda-1}$ and applies a non-linear
transformation to compute its own output $f_\lambda$. The behavior of these non-linearities
is controlled through a set of parameters $\theta_\lambda$, which are specific to each layer.
These parameters, also called \textit{weights}, link the neurons of a
given layer to the neurons of the layer that precedes it. They encode knowledge
extracted by the model from the training data (see below).
Hence, given an input $x$, a neural network $f$ performs the following computation
to predict its class:
\begin{equation}
\label{eq:dnn}
f(\theta, x) = f_{l-1}\left(\theta_{l-1}, f_{l-2}\left(\theta_{l-2}, ... f_0\left(\theta_0, x\right)\right)\right)
\end{equation}
When possible, we simplify the notation by omitting the vector parameters $\theta=[\theta_0, ..., \theta_{l-1}]$, in which
case we write $f(x)$.

During \textbf{training}, the model is presented with a large collection of known
input-output pairs $(x,y)\in (X,Y)$. To begin, initial values for the weights $\theta$ are drawn randomly.
We then take a \textit{forward pass} through the model: 
given an input $x\in X$ and label $y\in Y$, we compute its current belief $f(x)$, which
is a vector whose components are the estimated probability of $x$ belonging to each
of the classes: e.g., the 5-th component $f(x)[5]$ is $P(y=5\mid \theta, x)$.
The model's prediction error is estimated by computing the value of a \textit{cost}---or \textit{loss}---function
given the current prediction $f(x)$ and the true label $y$.
In the backward pass, this error is differentiated with respect to all of the parameters in $\theta$,
and their values are updated to improve the predictions of neural network $f$.
By iteratively taking forward and backward passes, values of the model
parameters that (approximately) minimize the loss function on the training
data are found.

The model is then deployed to predict on data unseen during training. 
This is the \textbf{inference} phase: the model takes a 
forward pass on the new---test---input and outputs a label prediction. One hopes that the model will \textit{generalize}
to this test data to infer the true label from the patterns it has
encountered in its training data. This is often but not always the case 
however, as evidenced by adversarial examples (see Section~\ref{ssec:back-robustness} below).

\section{On Confidence, Interpretability \& Robustness}
\label{sec:conf-inter-robustness}

We 
systematize knowledge from previous efforts that tackled the problem
of confidence in machine learning. Their strengths and limitations motivate the 
choices made in Section~\ref{sec:approach} to design our approach
for measuring confidence in DNNs.
Finally, we position our work among existing literature on interpretability
and robustness in (deep) machine learning.

\subsection{Confidence in Machine Learning}
\label{ssec:background-confidence}

There exist several sources of uncertainty in ML applications~\cite{kendall2017uncertainties}.
Observations made to collect the dataset introduce \textit{aleatoric uncertainty} by
not including all necessary explanatory variables in the
data. For instance, a spam dataset that only contains 
email metadata but not their content would introduce 
substantial aleatoric uncertainty. In this paper, we focus
on \textit{epistemic uncertainty}---or \textit{model uncertainty}---introduced 
by a ML model
because it is learned from limited data.

Below, we survey approaches for estimating the
confidence of DNNs. 
The most frequently used, adding a softmax layer to the 
DNN, is not reliable 
for inputs that
fall off the model's training manifold. Other approaches
like Bayesian deep learning remain computationally expensive. 
In Section~\ref{sec:approach}, we thus introduce an
approach  to provide more
reliable model uncertainty estimates.

\paragraph{Softmax probabilities} 
The output of DNNs used for classification
is a vector $f(x)$
that is typically interpreted as estimates 
of the model's confidence for classifying input $x$ in
class $j$ of the task considered.
This vector is almost always obtained
by adding a softmax layer that processes
the logits---or class scores---output by the penultimate model layer. 
Known as Platt scaling~\cite{platt1999probabilistic},
the approach is equivalent to fitting a logistic regression to the classifier's class scores:
\vspace*{-0.08in}
\begin{equation}
\label{eq:softmax}
f_{l-1}(x)[j] = \frac{e^{z_j}}{\sum_{p=0}^{n-1}z_p}
\vspace*{-0.07in}
\end{equation}
The logits $z_j=f_{l-2}\cdot w[j]$ (i.e., class scores) are originally floating point values
whose values are unbounded, whereas the output of the softmax is a vector
of floating point values that sum up to $1$ and
 are individually bounded between $0$ and $1$.

Contrary to
the popular belief, this approach is not a reliable estimator of confidence~\cite{gal2016uncertainty}.
Adversarial examples are good counter-examples to illustrate the weakness
of this metric: they are inputs crafted by adding a  perturbation
that force ML models to misclassify inputs that were originally
correctly classified~\cite{biggio2013evasion,szegedy2013intriguing}.
However, DNNs output a larger confidence for the wrong class
when presented with an adversarial example than the confidence they assigned
to the correct class when presented with a legitimate input~\cite{szegedy2013intriguing}:
In other words, the softmax indicates that
the DNN is more confident when it is 
mistaken than when its predicts the correct answer.

\paragraph{Bayesian approaches} 
Another popular class of techniques estimate the model uncertainty
that deep neural network architectures introduce because they
are learned from limited data
by involving the Bayesian formalism.
\textit{Bayesian deep learning} introduces a distribution over
models or their parameters (e.g., the weights that link different layers
of a DNN)  in order to offer principled uncertainty
estimates.
Unfortunately,
Bayesian inference remains computationally hard for neural networks.
Hence, different degrees of approximations are made to reduce the computational
overhead~\cite{mackay1992bayesian,graves2011practical,neal2012bayesian,blundell2015weight}, including about the prior
that is specified for the parameters of the neural network. Despite these
efforts, it remains difficult to implement Bayesian neural networks; thus
 radically different directions have been proposed recently. They require less
modifications to the learning algorithm. One such proposal is to use dropout at
test time to estimate uncertainty~\cite{gal2016dropout}. 
Dropout was originally introduced as a regularization technique
for training deep neural networks: for each forward and backward pass
pairs (see above for details), the output of 
a random subset of neurons is set to $0$: i.e., they are \textit{dropped}~\cite{srivastava2014dropout}. 
Gal et al. instead  proposed to use dropout at test time and
cast it as approximate Bayesian inference~\cite{gal2016dropout}.
Because dropout can
be seen as an ensembling method, this approach naturally generalizes to using ensembles of models
as a proxy to estimate predictive uncertainty~\cite{lakshminarayanan2017simple}.

\subsection{Interpretability in Machine Learning} 

It is not only difficult to calibrate DNN predictions to obtain
reliable confidence estimates, but also to present
a human observer with an explanation for model outputs.
Answering the question ``Why did the model decide that?'' can be a complex
endeavor for DNNs when compared to somewhat more interpretable models like
decision trees (at least trees that are small enough for a human
to understand their decision process). The ability to explain the logic followed by a model is also
key to debug ML-driven autonomy.\todo{Patrick: DARPA example?}

Progress in this area of \textit{interpretability} for ML (or sometimes explainable AI), remains limited because the criteria for success are ill-defined
and difficult to quantify~\cite{lipton2016mythos}.
Nevertheless, legislation like the European Union's General Data Protection
Regulation require that companies deploying ML and other forms of analysis on certain
sensitive data provide such interpretable outputs if the 
prediction made by a ML model is used to make decisions
without a human in the loop~\cite{goodman2016european}; 
and as a result,
there is a growing body of literature addressing interpretability~\cite{erhan2009visualizing,kim2015mind,ribeiro2016should,alain2016understanding,ribeiro2016should,kim2017tcav,bau2017network,dabkowski2017real}.

A by-product of the approach introduced in Section~\ref{sec:approach} 
is that it returns \textit{exemplar
inputs}, also called prototypes~\cite{bien2011prototype,li2017deep}, to 
interpret predictions through training
points that best explain the model's output because they are
processed similarly to the test input considered. 
This approach to interpretability through an \textit{explanation by example}
was pioneered by Caruena et al.~\cite{caruana1999case}, who suggested
that a comparison of the representation predicted by
a single layer neural network with the representations learned 
on its training data would help identify points in the training
data that best explain the prediction made. Among notable
follow-ups~\cite{kim2014bayesian,doshi2015graph,hendricks2016generating}, this technique
was also applied to visualize relationships learned between
words by the word2vec language model~\cite{mikolov2013distributed}.
As detailed in
Section~\ref{sec:approach}, we search for nearest neighboring
training points not at the level of the embedding layer only but
at the level of each layer within the DNN and use the labels of
the nearest neighboring training points to provide confidence, 
interpretability and robustness.

Evaluating interpretability is difficult because of the
involvement of humans.
Doshi-Velez and Kim~\cite{doshi2017towards} identify two classes of
evaluation for interpretability: (1)
the model is useful for a practical (and perhaps simplified) 
application used as a proxy to
test interpretability~\cite{ribeiro2016should,lei2016rationalizing,kim2015mind}
or (2) the model is learned using a specific hypothesis class
already established to be interpretable 
(e.g., a sparse linear model or a decision 
tree)~\cite{lou2012intelligible}. Our evaluation falls under
the first category: exemplars returned 
by our model simplify downstream practical applications.

\subsection{Robustness in Machine Learning}
\label{ssec:back-robustness}

The lack of confidence and interpretability of DNN outputs is also illustrated
by \textit{adversarial examples}: models make mistake on
these malicious inputs, yet their confidence is often higher in the mistake than when predicting on a legitimate input.

Adversarial examples are malicious test inputs obtained by perturbing  legitimate inputs~\cite{szegedy2013intriguing,biggio2013evasion,goodfellow2014explaining,papernot2016limitations,liu2016delving,sharif2016accessorize,hayes2017machine,carlini2017towards,sitawarin2018rogue}.
Despite the inputs being originally correctly classified,
the perturbation added to craft an adversarial example
changes the output of a model. 
In computer vision applications,
because the perturbation
added is so small in the pixel space, humans are typically unaffected: adversarial images are visually indistinguishable~\cite{szegedy2013intriguing}.
This not only shows that ML models lack \textit{robustness} to 
perturbations of their inputs, but also again that their predictions lack
human interpretability.

Learning models robust to adversarial examples is a challenging
task. Defining robustness is difficult, and the community has resorted
to optimizing for robustness in a $\ell_p$ norm ball around the training
data (i.e., making sure that the model's predictions are constant in a
neighborhood of each training point defined using a $l_p$ norm).
Progress has been made by discretizing the input space~\cite{anonymous2018thermometer},
training on adversarial examples~\cite{madry2017towards,tramer2017ensemble}
or attempting to remove the adversarial perturbation~\cite{xu2017feature,meng2017magnet}.
Using robust optimization would be ideal but is difficult---often
intractable---in practice because rigorous definitions
of the input domain region that adversarial examples make up are
often non-convex and thus difficult to optimize for.
Recent investigations showed that a convex region that
includes this non-convex region defined by adversarial examples 
can be used to upper bound the potential loss inflicted by 
adversaries and thus perform robust optimization over it. Specifically,
robust optimization is performed over a convex 
polytope that includes the non-convex space of adversarial
examples~\cite{kolter2017provable,raghunathan2018certified}.

Following the analysis of Szegedy et al.~\cite{szegedy2013intriguing}
suggesting that each layer of a neural network has large Lipschitz 
constants, there has been several attempts at making the representations
better behaved, i.e., to 
prove small Lipschitz constants per layer, which
would imply robustness to adversarial perturbations:
small changes to the input of a layer would be guaranteed
to produce bounded changes to the output of said layer. 
However, techniques proposed have either restricted the ability to train a neural 
network (e.g., RBF units~\cite{goodfellow2014explaining}) or 
demonstrated marginal improvements in robustness (e.g., 
Parseval networks~\cite{cisse2017parseval}). The 
approach introduced in Section~\ref{sec:approach}
instead uses a nearest neighbors operation to ensure  
representations output by layers at test time are consistent
with those learned during training. 
 \section{Deep k-Nearest Neighbors Algorithm}
\label{sec:approach}
The approach we introduce takes a DNN trained using
any standard DNN learning algorithm
and modifies the procedure followed to have the model predict on test data:
patterns identified in the data at test time by internal components (i.e., layers) 
of the DNN are
compared to those found during training
to ensure that any prediction made is supported by the training data.
Hence, rather than treating the model as a black-box and trusting its
predictions obliviously, our inference
procedures ensures that each intermediate computation performed
by the DNN is consistent with its final output---the
label prediction.

The pseudo-code for our Deep k-Nearest Neighbors (DkNN) procedure
is presented in Algorithm~\ref{alg:dknn}. We first motivate why 
analyzing representations internal to the deep neural network (DNN)
that underlies it allows the DkNN algorithm to strengthen the 
interpretability and robustness of its predictions. This is the object of 
Section~\ref{ssec:approach-representations}. Then, in Section~\ref{ssec:approach-conformal}, we inscribe our algorithm in the
framework of conformal prediction to estimate and calibrate the confidence of 
DkNN predictions. The confidence, interpretability and robustness
of the DkNN are empirically evaluated respectively in Sections~\ref{sec:eval-confidence},~\ref{sec:eval-interpretability} and~\ref{sec:eval-robustness}.

\subsection{Predicting with Neighboring Representations}
\label{ssec:approach-representations}

\subsubsection{Motivation}

As we described in Section~\ref{sec:background}, DNNs learn a hierarchical set of representations. In other
words, they project the input in increasingly abstract spaces,
and eventually in a space where the classification decision can
be made using a logistic regression---which is the role of
the softmax layer typically used as the last layer of neural
network classifiers (see Section~\ref{ssec:background-confidence}). 
In many cases, this hierarchy of representations enables DNNs to generalize
well  on data presented to the
model at test time. However, phenomena like adversarial examples---especially those produced by feature adversaries~\cite{sabour2015adversarial}, which we cover
extensively in Section~\ref{ssec:dknn-robust-feature-adv}---or
the lack of invariance to translations~\cite{engstrom2017rotation} indicate that
representations learned by DNNs are not as robust as the technical community initially
expected. 
Because DNN training algorithms make the implicit assumption that
test data is drawn from the same distribution than training
data, this has not been an obstacle to most developments of ML.
However, when one wishes to deploy ML in settings where
safety or security are critical, it becomes necessary to invent
mechanisms suitable to identify when the model is extrapolating too
much from the representations it has built with its training data.
Hence, the first component of our approach analyzes these
internal representations at test time to detect inconsistencies
with patterns analyzed in the training data.

Briefly put, our goal is to ensure that each intermediate computation
performed by the deep neural network is conformal with the final
prediction it makes. Naturally, we rely on the layered structure afforded
by DNNs to define these intermediate computation checks. Indeed,
each hidden layer, internal to the DNN, outputs a different
representation of the input presented to the model. Each layer
builds on the representation output by the  layer that precedes it
to compute its own representation
of the input. When the final layer of a DNN indicates that the input
should be classified 
in a particular class, it outputs in a way an abstract representation
of the input which is the class itself. This representation is
computed based on the representation output by the penultimate layer,
which itself must have been characteristic of inputs from this
particular class. The same reasoning can be recursively applied to
all layers of the DNN until one reaches the input layer, i.e., the
input itself.

In the event where a DNN mistakenly predicts the wrong class 
for an input, there is necessarily one of its layers that transformed
the input's representation, which was by definition in the correct 
class initially because the input is itself a representation in the
input domain, into a representation that is closer to inputs from the
wrong class eventually predicted by the DNN. This behavior,
depicted in Figure~\ref{fig:dknn-motivation}, is what our approach
sets out to algorithmically characterize to make DNN predictions more confident,
interpretable and robust.

\subsubsection{A nearest neighbors approach} In its simplest form,
our approach (see Figure~\ref{fig:dknn-motivation}) can be understood as creating a nearest neighbors
classifiers in the space defined by each DNN layer. While prior work
has considered fitting linear classifiers~\cite{alain2016understanding} or support vector machines~\cite{lee2015deeply},
we chose the nearest neighbors because it explicits the relationship
between predictions made by the model and its training data. We later leverage
this aspect to characterize the nonconformity of model predictions.

\begin{algorithm}[t]
	\caption{\textbf{-- Deep k-Nearest Neighbor}.}
	\label{alg:dknn}
	\begin{algorithmic}[1] 
		\Require training data $(X, Y)$, calibration data $(X^c, Y^c)$
		\Require trained neural network $f$ with $l$ layers
		\Require number $k$ of neighbors
		\Require test input $z$
		\State // Compute layer-wise $k$ nearest neighbors for test input $z$
		\For{each layer $\lambda \in 1 \text{ .. } l$}
		\State $\Gamma \leftarrow$ $k$ points in $X$ closest to $z$ found w/ LSH tables 
		\State $\Omega_\lambda \leftarrow \{Y_i : i \in \Gamma\} $ \Comment{Labels of $k$ inputs found}
		\EndFor
		\State // Compute prediction, confidence and credibility 
		\State $A=\{\alpha(x,y) : (x, y) \in (X^c, Y^c)\}$ \Comment{Calibration}
		\For{each label $j\in 1..n$}
		\State $\alpha(z, j) \leftarrow \sum_{\lambda \in 1 .. l} |i \in \Omega_\lambda : i \neq j |$ \Comment{Nonconformity}
		\State $p_j(z) = \frac{\left| \{ \alpha \in A : \alpha \geq \alpha(z, j) \} \right|}{|A|}$ \Comment{empirical $p$-value}
		\EndFor
		\State prediction $\leftarrow \arg\max_{j\in 1..n} p_j(z)$
		\State confidence $\leftarrow 1 - \max_{j\in 1..n, j\neq prediction} p_j(z)$
		\State credibility $\leftarrow \max_{j\in 1..n} p_j(z)$
		\State \Return prediction, confidence, credibility
	\end{algorithmic}
\end{algorithm}

Once the neural network $f$ has been trained, we record the
output of its layers $f_\lambda$ for $\lambda\in 1..l$ 
on each of its training points. For each layer $\lambda$, we thus have
a representation of the training data along with the corresponding labels,
which allows us to construct a nearest neighbors classifier.
Because the output of layers is often high-dimensional (e.g., 
the first layers of DNNs considered in our experiments
have tens of thousands of neuron activations),
we use the algorithm
of Andoni et al.~\cite{andoni2015practical} to efficiently perform  the lookup
of nearest neighboring representations in the high-dimensional spaces learned by DNN layers. It implements data-dependent
locality-sensitive hashing~\cite{gionis1999similarity,andoni2006near} to find nearest neighbors according to the cosine similarity
between vectors. Locality-sensitive hashing (LSH) functions differ
from cryptographic hash functions; they are designed to \textit{maximize}
the collision of similar items. This property is beneficial to the
nearest neighbors problem in high dimensions. Given a query point, 
locality-sensitive hashing functions are first used to establish a list 
of candidate nearest neighbors: these are points that collide (i.e., are similar) with
the query point. Then, the nearest neighbors can be found among
this set of candidate points.
In short, when we are presented with a test input $x$:
\begin{enumerate}
	\item We run input $x$ through the
	DNN $f$ to obtain the $l$ representations output by its layers: $\{f_\lambda(x)\mid \lambda\in 1 .. l \}$
	\item For each of these representations $f_\lambda(x)$, we use a
	nearest neighbors classifier based on locality-sensitive
	hashing to find the $k$
	training points whose representations at layer $\lambda$ are closest
	to the one of the test input (i.e., $f_\lambda(x)$).
	\item For each layer $\lambda$, we collect the multi-set $\Omega_\lambda$ of labels
	assigned in the training dataset to the $k$ nearest representations found at the previous step.
	\item We use all multi-sets $\Omega_\lambda$ to compute the prediction of our
	DkNN according to the framework of conformal prediction (see Section~\ref{ssec:approach-conformal}).
\end{enumerate}

The comparison of representations predicted by the DNN at test 
time with neighboring representations learned during training
allows us to make progress towards providing certain desirable
properties for DNNs such as interpretability and robustness. Indeed,
we demonstrate in Section~\ref{sec:eval-interpretability} that
the nearest neighbors offer a form of natural---and most importantly
human interpretable---explanations for the intermediate computations
performed by the DNN at each of its layers. Furthermore,
in order to manipulate the predictions of our DkNN algorithm with
malicious inputs like adversarial examples, adversaries have to force
inputs to closely align with representations learned from the
training data by all layers of the underlying DNN. Because the 
first layers learn low-level features and the adversary
is no longer able to exploit non-linearities to gradually change
the representation of an input from the correct class to the wrong
class, it becomes harder to produce adversarial examples with
 perturbations that don't change the semantics (and label)
 of the input. We validate these
claims in Section~\ref{sec:eval-robustness} but first
present the second component of our approach, from which 
stem our predictions and their calibrated confidence estimates.

\subsection{Conformal Predictions for DkNN Confidence Estimation}
\label{ssec:approach-conformal}

Estimating DNN confidence is difficult, and often an obstacle
to their deployment (e.g., in medical applications or security-critical
settings). Literature surveyed in Section~\ref{sec:background} and our experience
concludes that  probabilities
output by the softmax layer and commonly used as a proxy for
DNN confidence are not well calibrated~\cite{guo2017calibration}. In particular, they often overestimate
the model's confidence when making predictions on inputs that fall
outside the training distribution (see Section~\ref{ssec:out-of-distribution} for
an empirical evaluation). Here, we leverage ideas from
conformal prediction and the nearest neighboring representations
central to the DkNN algorithm to define how it makes predictions
accompanied with \textit{confidence} and \textit{credibility}
estimates. 
While confidence indicates how likely the prediction is to be correct
according to the model's training set, credibility quantifies how relevant the
training set is to make this prediction.
 Later, we demonstrate experimentally
that credibility is well calibrated when the DkNN predicts in both benign (Section~\ref{sec:eval-confidence}) and adversarial
(Section~\ref{sec:eval-robustness})
environments.

\subsubsection{Inductive Conformal Prediction} 
Conformal prediction builds on an existing ML classifier
to provide a probabilistically valid measure of confidence and credibility for
predictions made by the underlying classifier~\cite{saunders1999transduction,vovk1999machine,shafer2008tutorial}.
In its original variant, which we don't describe
here in the interest of space, conformal prediction required that 
the underlying classifier be trained from scratch for each test input. 
This cost would be prohibitive in our case, because the underlying ML classifier is a DNN. Thus, 
we use an \textit{inductive} variant of conformal 
prediction~\cite{papadopoulos2002inductive,papadopoulos2008inductive}
that does not 
require retraining the underlying classifier for each query because it assumes the 
existence of a calibration set---holdout data that does not overlap
with the training or test data.

Essential to all forms of, including inductive, conformal prediction is the existence of 
a \textit{nonconformity measure}, which
indicates how different a labeled input is from previous observations of samples from
the data distribution. Nonconformity is typically measured with the underlying
machine learning classifier. In our case, we would like to measure how different
a test input with a candidate label is from previous labeled inputs that make
up the training data. In essence, this is one of the motivations of the DkNN.
For this reason, a natural proxy for the nonconformity of a labeled input is
the number of nearest neighboring representations found by the DkNN in its training data 
whose label is \textit{different} from the candidate label. When this number is low, 
there is stronger support for the candidate label assigned to the test input 
in the training data modeled by the DkNN. Instead, when 
this number is high, the DkNN was not able to find training points that 
support the candidate label, which is likely to be wrong.
In our
case, nonconformity of an input $x$ with the label $j$ is defined as:
\begin{equation}
\label{eq:nonconformity}
\alpha(x, j) = \sum_{\lambda \in 1 .. l} |i \in \Omega_\lambda : i \neq j |
\end{equation}
where $\Omega_\lambda$ is the multi-set of labels for the training points whose
representations are closest to the test input's at layer $\lambda$.

Before inference can begin and predictions made on unlabeled test inputs, 
we compute the nonconformity of the calibration dataset $(X^c, Y^c)$, which is labeled. The calibration data is sampled from the same distribution than
 training data but is not used to train the model. 
The 
size of this calibration set should strike a balance between reducing the number of
points that need to be heldout from training or test datasets and increasing the
precision of empirical $p$-values computed (see below)
Let us denote with $A$ the nonconformity values computed on the set of
calibration data, that is $A=\{\alpha(x,y) : (x, y) \in (X^c, Y^c)\}$.  

Once all of these values are
computed, the nonconformity score
of a test input is compared with the scores computed on the calibration dataset 
through a form of hypothesis testing.
Specifically, given a test input $x$, we perform the following
for each candidate label $j\in 1..n$:
\begin{enumerate}
	\item We 
	use neighbors identified by the DkNN and Equation~\ref{eq:nonconformity}
	to compute the nonconformity $\alpha(x, j)$ where $x$ is the test input
	and $j$ the candidate label.
	\item We calculate the fraction of nonconformity measures for the calibration data that
	are larger than the test input's. This is the empirical $p$-value
	of candidate label $j$:
	\begin{equation}
	p_j(x) = \frac{\left| \{ \alpha \in A : \alpha \geq \alpha(x, j) \} \right|}{|A|}
	\end{equation}
\end{enumerate}  
The \textit{predicted label} for the test input is the one assigned the largest empirical
$p$-value, i.e., $\arg\max_{j\in 1..n} p_j(x)$. Although two classes could be assigned identical empirical $p$-values, this does not happen often in practice. The prediction's \textit{confidence} 
is set to $1$ minus the second largest empirical $p$-value. Indeed, this quantity is the
probability that any label other than the prediction is the true label. 
Finally, the prediction's \textit{credibility}
is the empirical $p$-value of the prediction: it bounds the nonconformity of any label assigned
to the test input with the training data. In our experiments,
we are primarily interested in the second metric, credibility.

Overall, the approach described in this Section yields the inference procedure outline in
Algorithm~\ref{alg:dknn}.

 \section{Evaluation of the Confidence of DkNNs}
\label{sec:eval-confidence}

The DkNN leverages nearest neighboring representations
in the training data to define the nonconformity of individual predictions.
Recall that in Section~\ref{sec:approach}, we applied the framework of 
conformal prediction to define two notions of what commonly
falls under the realm of ``confidence'':
\textit{confidence} and \textit{credibility}. 
In our experiments, we measured high confidence on a large majority of inputs. In other words,
the nonconformity of the runner-up candidate label is high, making it
unlikely---according to the training data---that this second label be the true
answer. However, we observe that credibility varies  across both in-
and out-of-distribution samples. 

Because our primary interest is the support (or lack of thereof) that training data gives to 
DkNN predictions, which is precisely what credibility characterizes, we tailor our evaluation to
demonstrate that credibility is well-calibrated, i.e., identifies predictions not supported
by the training data. For instance, we validate the low
credibility of DkNN predictions on out-of-distribution inputs. 
Here, all experiments are conducted
in benign settings, whereas an evaluation of the DkNN
in adversarial settings is found later in
Section~\ref{sec:eval-robustness}.

\subsection{Experimental setup} We experiment with three datasets. First, the handwritten recognition
task of MNIST is a
classic ML benchmark: inputs are grayscale images of zip-code digits
written on postal mail and the task classes (i.e., the model's expected
output), digits from 0 to 9~\cite{lecun2010mnist}. Due to artifacts it possesses, e.g., redundant features, we use MNIST as a ``unit test'' to 
validate our DkNN implementation. Second, the SVHN dataset is another seminal
benchmark collected by Google Street View:  inputs
are colored images of house numbers and classes, digits from 0
to 9~\cite{netzer2011reading}. This task is harder than MNIST because inputs are not as
well pre-processed: e.g., images in the same class have different colors and light conditions.
Third, the GTSRB dataset
is a collection of traffic sign images to be classified
in 43 classes (each class corresponds to a type of
traffic sign)~\cite{Stallkamp2012}. For all datasets, we use a DNN
that stacks convolutionnal layers with
one or more fully connected layers. 
Specific architectures used for each of the three datasets
are detailed in the Appendix.

For each model, we implement the DkNN classifier  described
in Algorithm~\ref{alg:dknn}. We use a grid parameter search to set 
the number of neighbors $k=75$ and the  size (750 for MNIST
and SVHN, 850 for GTSRB because the task has more classes)
of the calibration set, which is
obtained by holding out a subset of the test data not used for evaluation. 
Table~\ref{tbl:classification-accuracy} offers a comparison of classification accuracies for the DNN and DkNN. The impact of the DkNN on performance
is minimal when it does not improve performance (i.e., increase accuracy).

\subsection{Credibility on in-distribution samples}

Given that there exists no ground-truth for the credibility of a
prediction, qualitative or quantitative evaluations of 
credibility estimates is often difficult. 
Indeed, datasets include the true label along with training and
test inputs but do not
indicate the expected credibility of a model  on these inputs. Furthermore, these inputs may for instance present some ambiguities
and as such make $100\%$ credibility a non-desirable outcome.

One way to characterize well-behaved credibility estimates is that
they should truthfully convey the likelihood of a prediction
being correct: high confident predictions should almost always
be correct and low confident predictions almost always wrong.
Hence, we plot \textit{reliability diagrams} to visualize the
calibration of our credibility~\cite{niculescu2005predicting}. They are histograms presenting
accuracy as a function of credibility estimates. Given
a set of test points $(X, Y)$, we group them in bins
$(X_{ab}, Y_{ab})$. A point $(x,y)$ is placed in bin $(X_{ab}, Y_{ab})$ if the model's
credibility on $x$ is contained in the interval $[a, b)$. Each bin is assigned the model's mean accuracy as its value:
\begin{equation}
\label{eq:reliability-diagrams}
bin(a, b) \coloneqq \frac{1}{|X_{ab}|} \sum_{(x, y)\in (X_{ab},Y_{ab})} 1_{f(x) = y}
\end{equation}
Ideally, the credibility of a well-calibrated model should increase
linearly with the accuracy of predictions, i.e., the reliability
diagram should approach a linear
function: $bin(a, b) \simeq \frac{a + b}{2}$.

\begin{table}[t]
	\centering
	{\renewcommand{\arraystretch}{1.2}
		\begin{tabular}{|c||c|c|}
			\hline
			 Dataset & DNN Accuracy & DkNN  Accuracy \\  \hline\hline
			 MNIST  & 99.2\% &  99.1\% \\\hline 
			SVHN  & 90.6\%  & 90.9\% \\\hline 
			GTSRB  & 93.4\%  & 93.6\% \\\hline
		\end{tabular}
		\vspace*{0.1in}
	}
	\caption{\textbf{Classification accuracy of the DNN and DkNN}: the DkNN has a limited impact on or improves performance.}
	\label{tbl:classification-accuracy}
\end{table}

\begin{figure}[t] 
	\begin{subfigure}[b]{0.5\columnwidth}
		\centering
		\includegraphics[width=\textwidth]{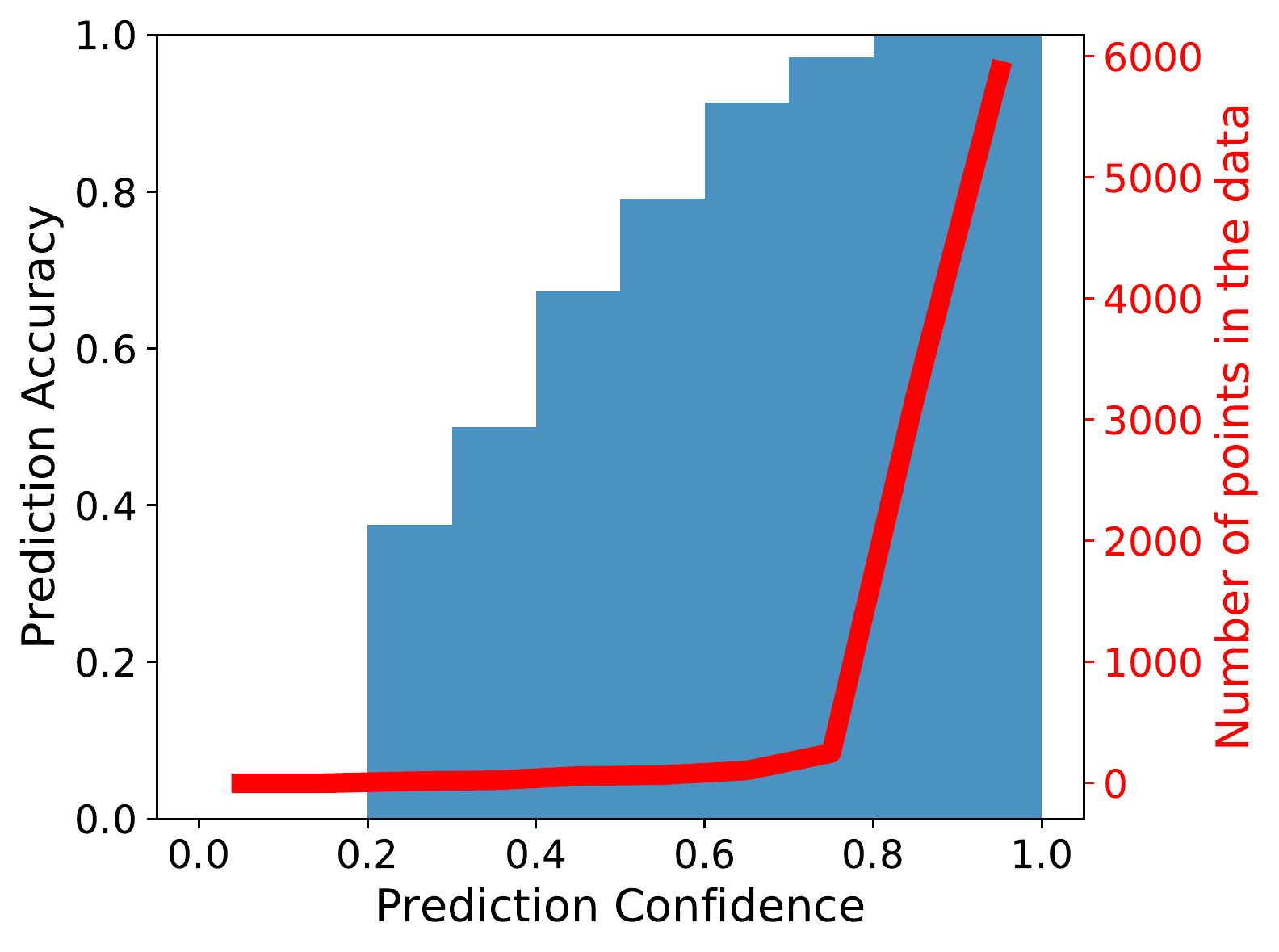} 
		\caption{Softmax - MNIST test set}
	\end{subfigure}~
	\begin{subfigure}[b]{0.5\columnwidth}
		\centering
		\includegraphics[width=\textwidth]{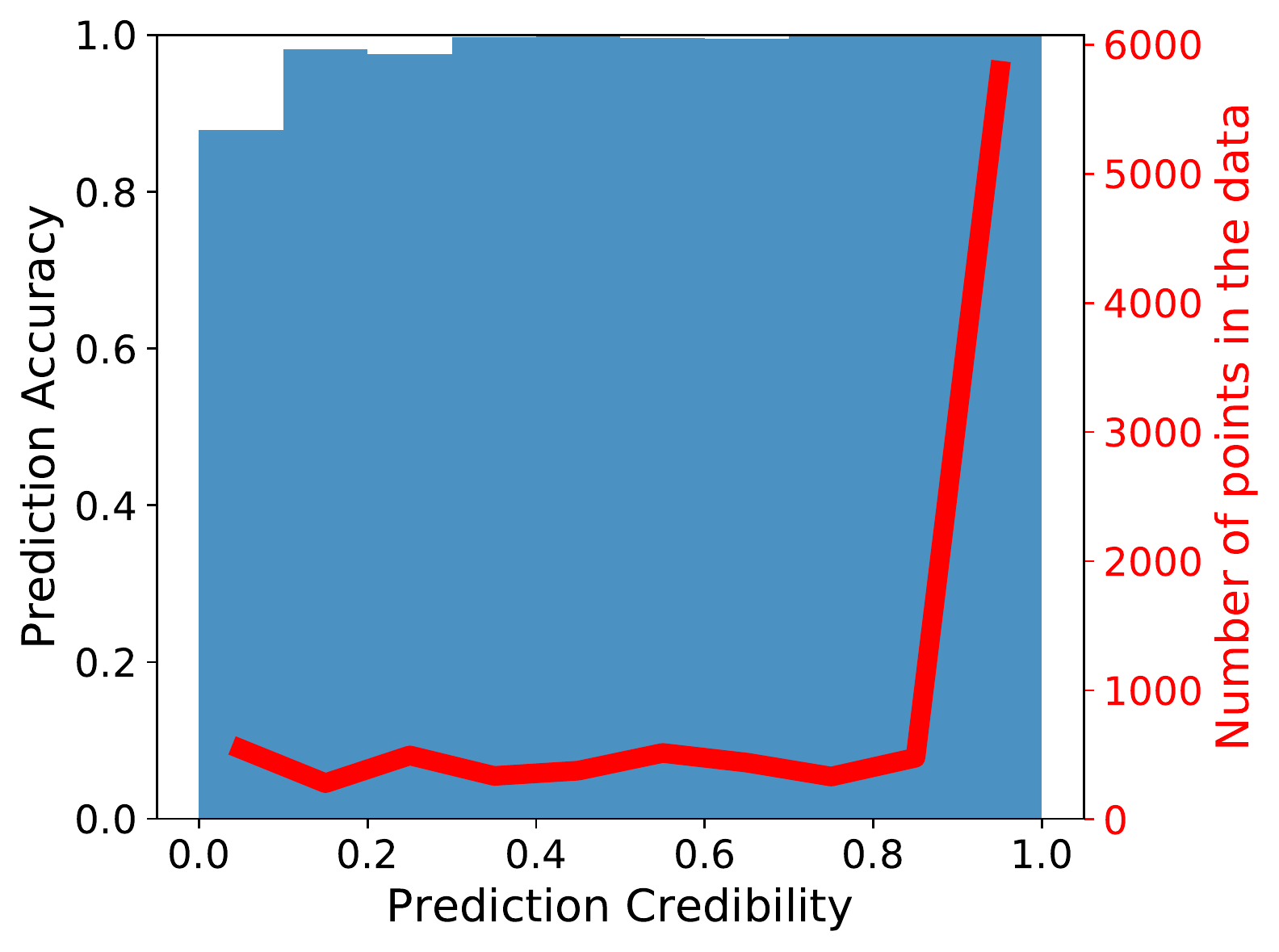} 
		\caption{DkNN - MNIST test set}
	\end{subfigure}
	\begin{subfigure}[b]{0.5\columnwidth}
		\centering
				\includegraphics[width=\textwidth]{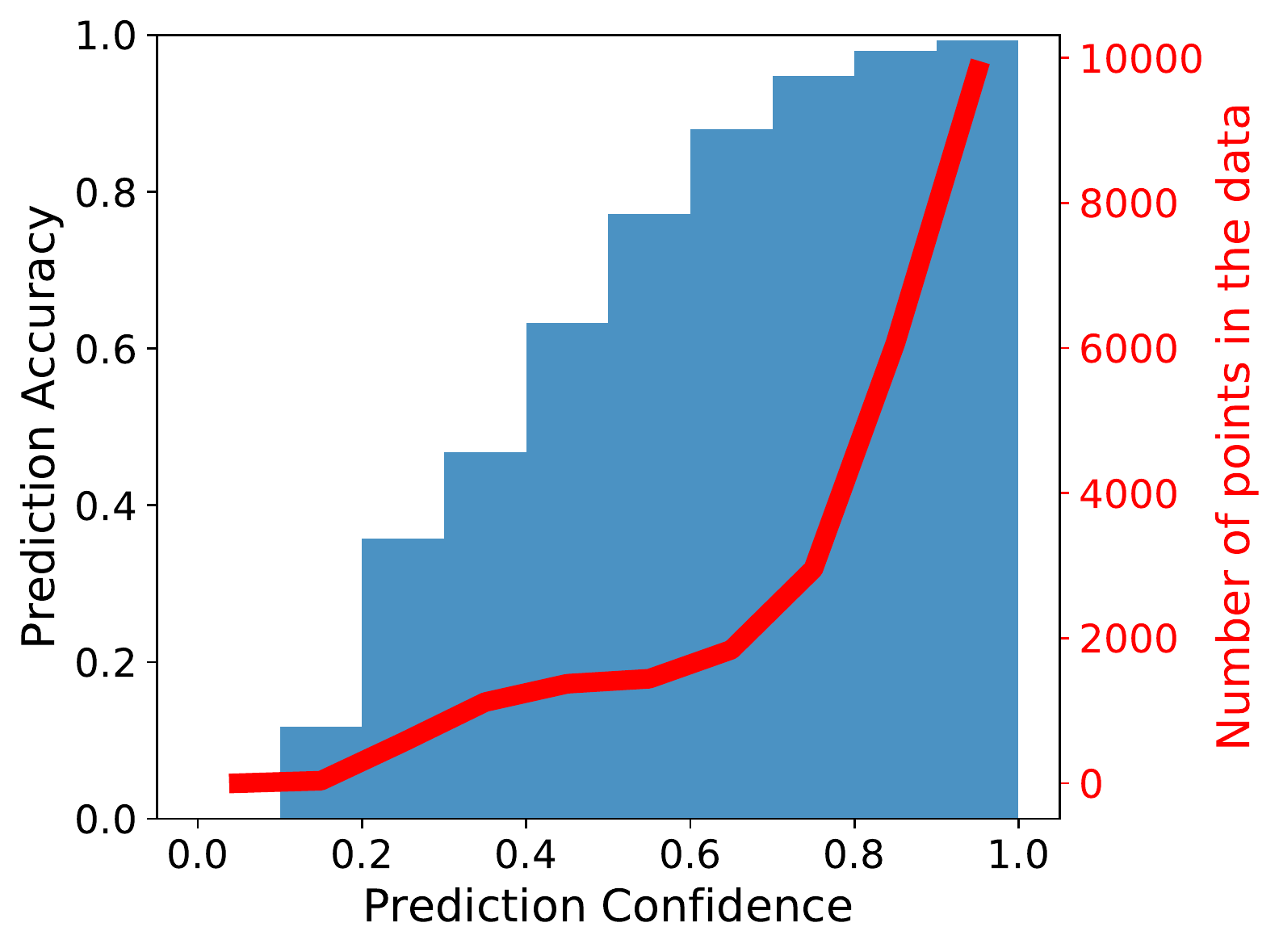} 
		\caption{Softmax - SVHN test set}
	\end{subfigure}~
	\begin{subfigure}[b]{0.5\columnwidth}
		\centering
				\includegraphics[width=\textwidth]{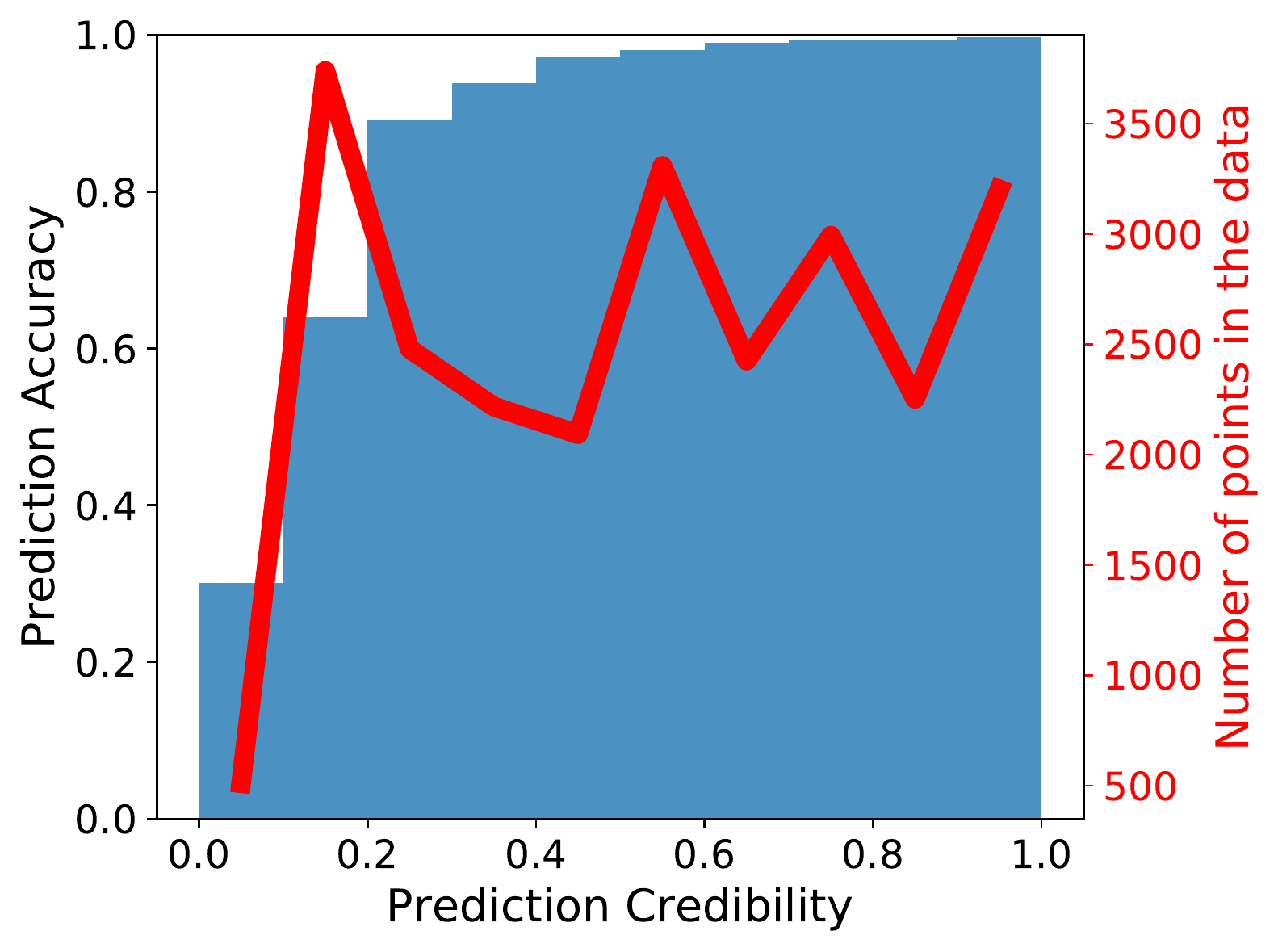} 
		\caption{DkNN - SVHN test set}
	\end{subfigure}
	\begin{subfigure}[b]{0.5\columnwidth}
		\centering
				\includegraphics[width=\textwidth]{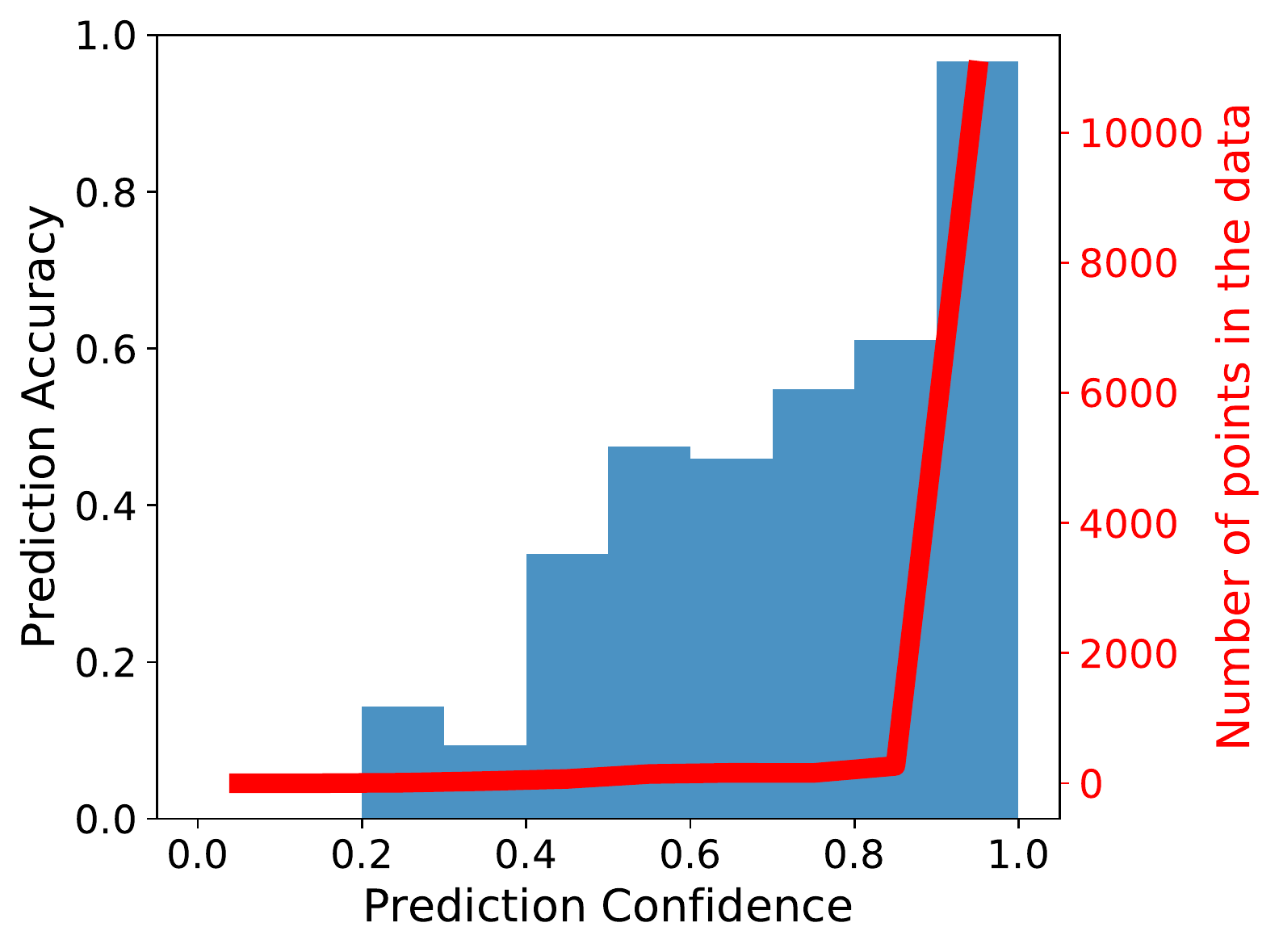} 
		\caption{Softmax - GTSRB test set}
	\end{subfigure}~
	\begin{subfigure}[b]{0.5\columnwidth}
		\centering
				\includegraphics[width=\textwidth]{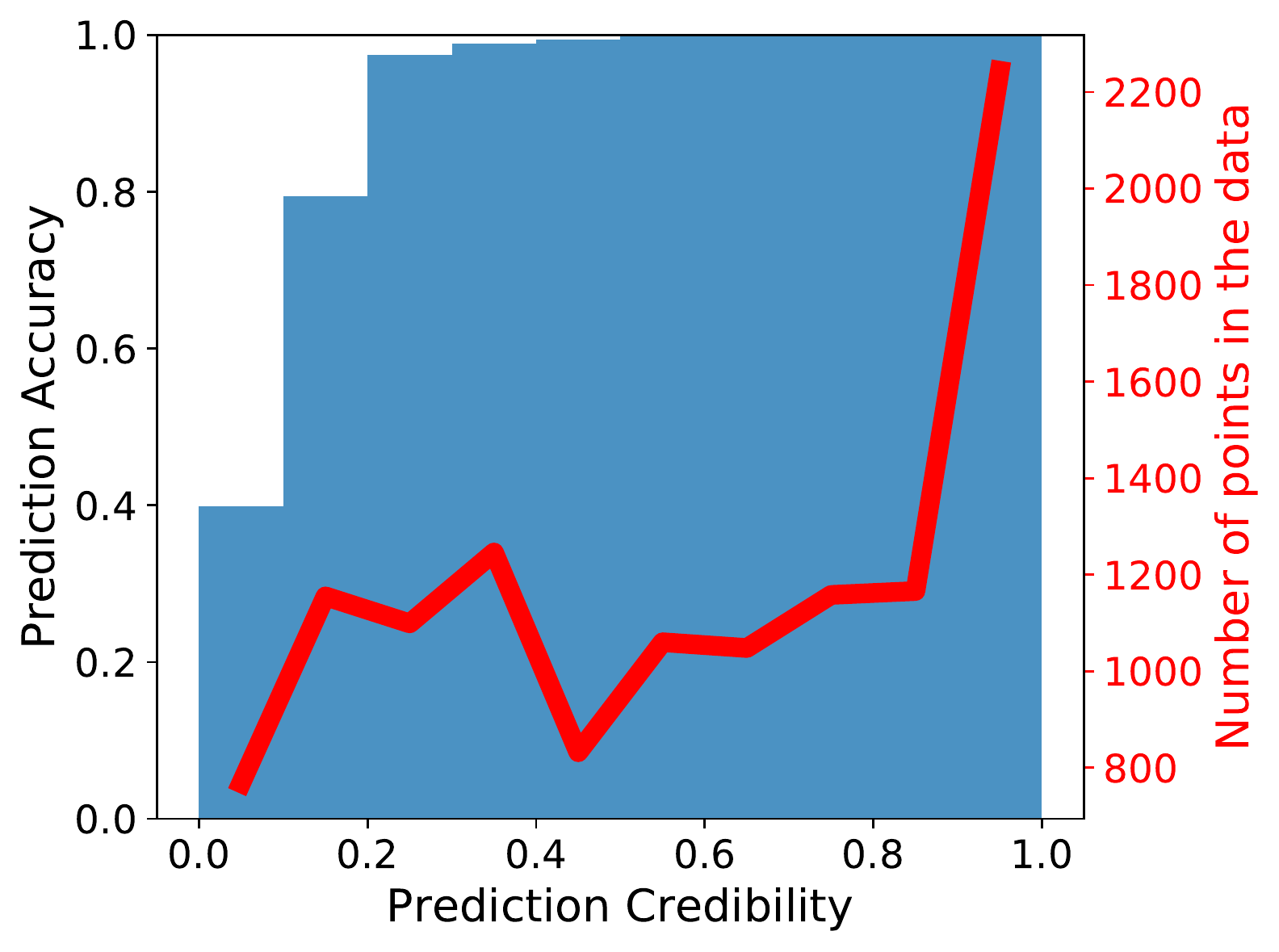} 
		\caption{DkNN - GTSRB test set}
	\end{subfigure}
	\caption{\textbf{Reliability diagrams of DNN softmax confidence (left) and DkNN credibility (right) on test data}---bars (left axis) indicate the mean accuracy of predictions binned by credibility; the red line (right axis) illustrates data density across bins. The softmax outputs 
	high confidence on most of the data while DkNN credibility 
	spreads across the value range.}
	\label{fig:dknn-relia-diag-clean} 
\end{figure}

Reliability diagrams are plotted for the MNIST, SVHN and GTSRB dataset in Figure~\ref{fig:dknn-relia-diag-clean}. On the left, they
visualize confidence estimates output by the DNN softmax;
that is the probability $\arg\max_j f_j(x)$ assigned to the most likely class.
On the right, they plot the credibility of DkNN predictions, as defined
in Section~\ref{sec:approach}.
At first, it may appear that the
softmax is better calibrated than its DkNN counterpart: its reliability diagrams 
are closer to the linear relation between accuracy and DNN confidence.
However, if one takes into account the distribution of DkNN credibility
values across the test data (i.e., the number of test points
found in each credibility bin reflected by the red line
overlaid on the bars), it surfaces that the 
softmax is almost always very confident on test data with
a confidence above $0.8$.
Instead, the DkNN  uses
the range of possible credibility values for datasets like SVHN,
whose test set contains a larger number of inputs that are
difficult to classify (reflected by the lower mean accuracy of the
underlying DNN). 
We will see how this
behavior is beneficial when processing 
out-of-distribution samples in Section~\ref{ssec:out-of-distribution} below
and adversarial examples later in Section~\ref{sec:eval-robustness}.

\begin{figure}[t] 
	\begin{subfigure}[b]{\columnwidth}
		\centering
		\includegraphics[width=0.7\columnwidth]{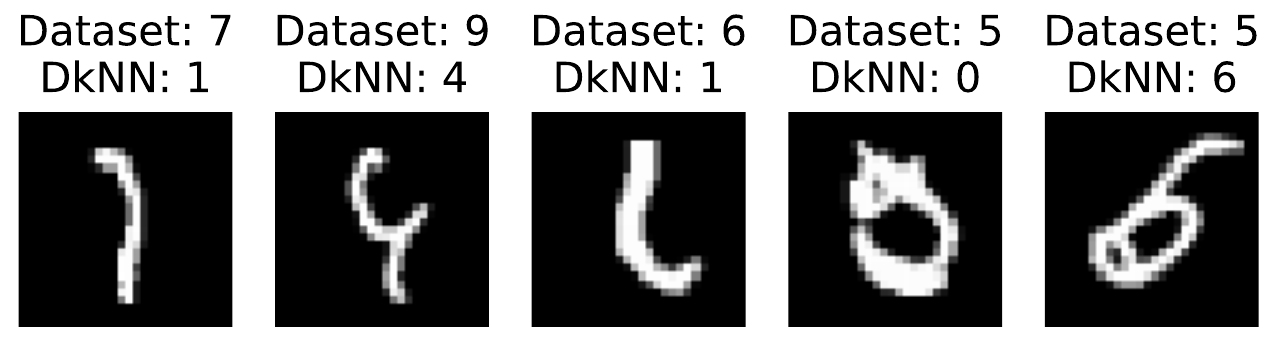} 
	\end{subfigure}
	\begin{subfigure}[b]{\columnwidth}
		\centering
		\includegraphics[width=0.7\columnwidth]{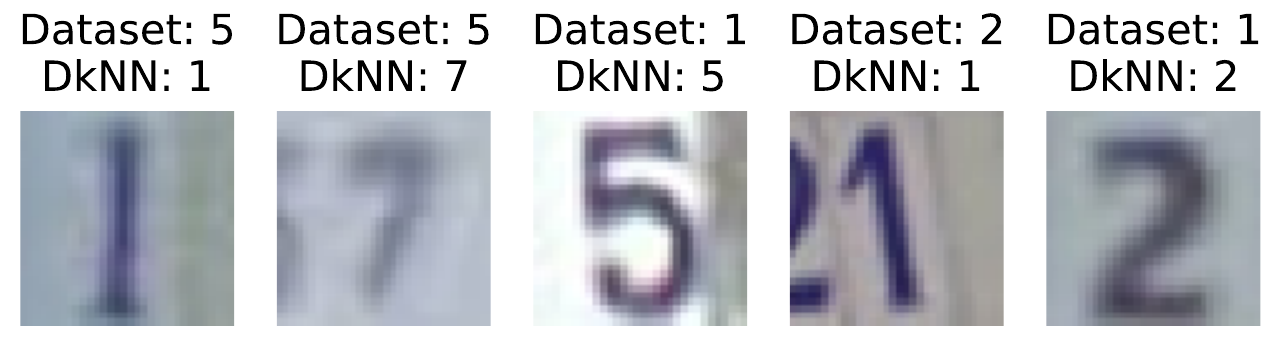} 
	\end{subfigure}
	\caption{\textbf{Mislabeled inputs from the MNIST (top) and SVHN (bottom) test sets}: we found these points by searching for inputs that are classified with strong credibility by the DkNN in a class that is different than the label found in the dataset. }
	\label{fig:dknn-misclassified-dataset-inputs} 
\end{figure}

In addition,the credibility output by the DkNN 
provides insights into the test data itself. 
For instance, we find that credibility  
is sufficiently reliable to find test inputs whose label \textit{in the original dataset} is
wrong. 
In both the MNIST and SVHN test sets, we looked for images that were 
were assigned a
high credibility estimate by the DkNN for a label that did not match
the one found in the dataset: i.e., the DkNN ``mispredicted''
the input's class according to the label included in the dataset. Figure~\ref{fig:dknn-misclassified-dataset-inputs} depicts some of the images
returned by this heuristic. It is clear for all of them that the dataset
label is either wrong (e.g., the MNIST 4 labeled as a 9 and SVHN 5 labeled
as a 1) or ambiguous (e.g., the penultimate MNIST image is likely to be
a 5 corrected into a 0 by the person who wrote it, and two of the SVHN
images were assigned the label of the digit that is cropped on the left of the
digit that is in the center). We were not able to find mislabeled inputs in
the GTSRB dataset.

\begin{figure}[t] 
	\begin{subfigure}[b]{0.5\columnwidth}
		\centering
		\includegraphics[width=\columnwidth]{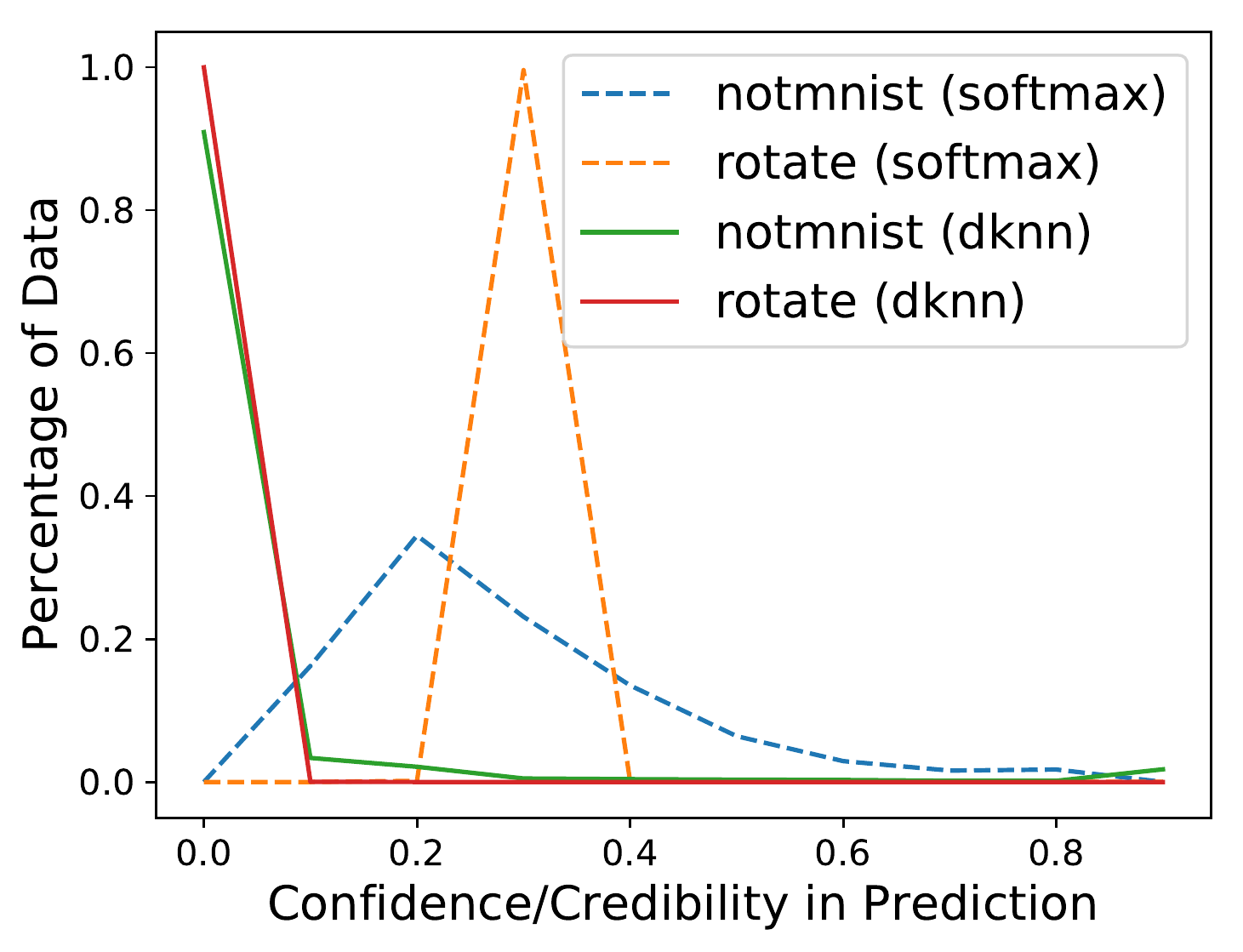} 
		\caption{MNIST}
	\end{subfigure}~
	\begin{subfigure}[b]{0.5\columnwidth}
		\centering
		\includegraphics[width=\columnwidth]{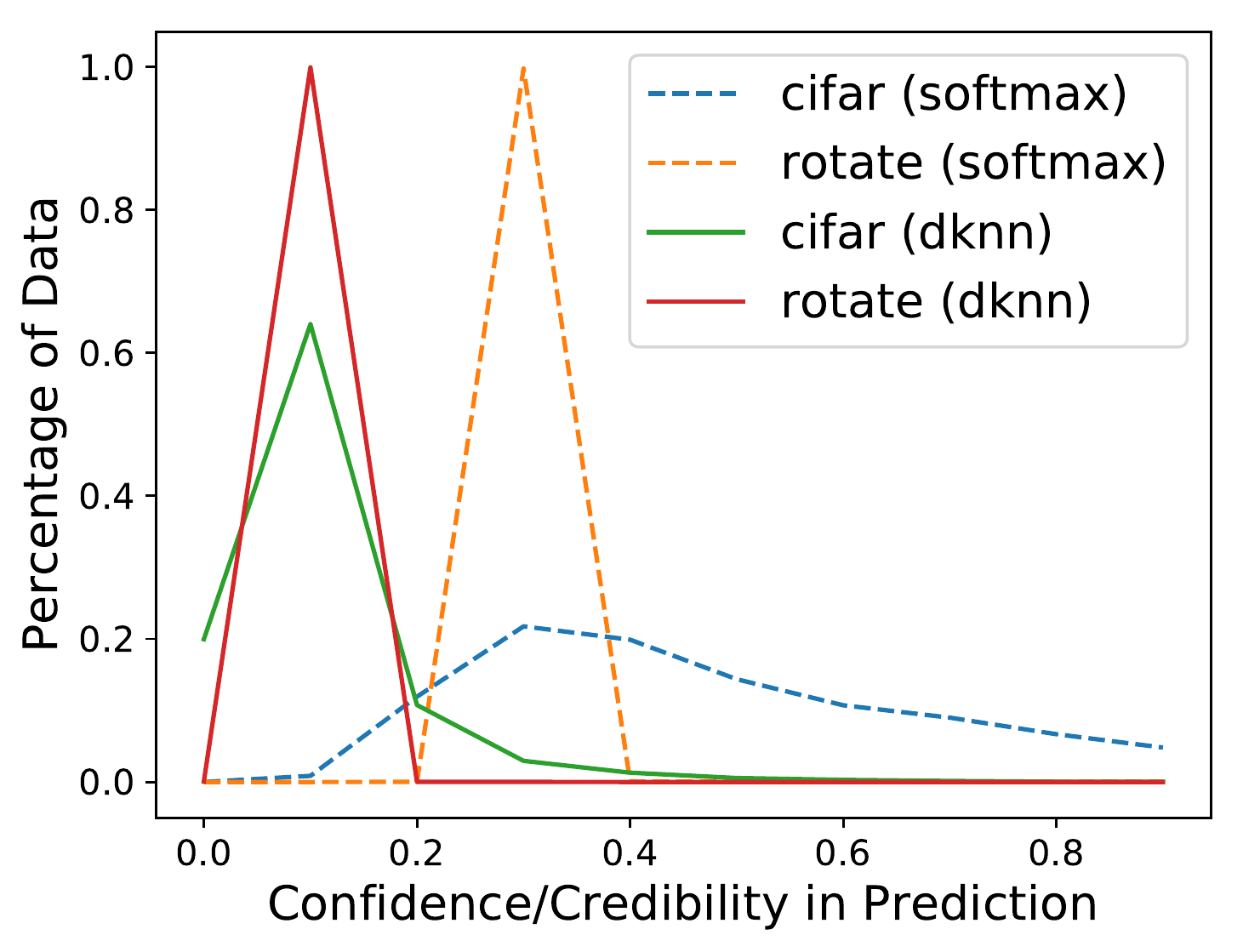} 
		\caption{SVHN}
	\end{subfigure}
	\caption{\textbf{DkNN credibility vs. softmax confidence on out-of-distribution test data}: the lower credibility of
	DkNN predictions (solid lines) compared to the softmax confidence (dotted lines) 
	is desirable here because test inputs are not part of the 
	distribution on which the model was trained---they are
	from another dataset or created by rotating inputs.}
	\label{fig:out-of-distribution-confidence} 
\end{figure}

\subsection{Credibility on out-of-distribution samples}
\label{ssec:out-of-distribution}

We now validate the DkNN's prediction credibility
on out-of-distribution samples. Inputs considered in this
experiment are either from a different classification task (i.e.,
drawn from another distribution) or generated by
applying geometrical transformations to inputs sampled from the distribution.
Due to the absence of support for these inputs in our training
manifold, we expect the DkNN's credibility  to be
low on these inputs: the training data used to learn the model
is not relevant to classify the test inputs we ask the model to classify.

For MNIST, the first set of out-of-distribution samples
contains images from the NotMNIST dataset,
which are images of unicode characters rendered using computer
fonts~\cite{bulatov2011notmnist}. Images from NotMNIST have an identical format to MNIST 
but the classes are non-overlapping: none of the classes
from MNIST (digits from 0 to 9) are included in the NotMNIST dataset 
(letters from A to J) and vice-versa. 
For SVHN, the analog set of out-of-distribution samples 
contains images from the CIFAR-10 
dataset: they have the same format but again 
there is no overlap between 
SVHN and the objects and animals from CIFAR-10~\cite{krizhevsky2009learning}.
For both the MNIST and SVHN datasets, we rotate all test inputs by
an angle of \ang{45}
to generate a
second set of out-of-distribution samples. Indeed, despite the presence of convolutional layers
to encourage invariance to geometrical transformations, 
DNNs poorly classify rotated data unless they
are explicitly trained on examples of such rotated inputs~\cite{engstrom2017rotation}.

The credibility of the DkNN
 on these out-of-distribution samples is compared with the probabilities
predicted by the underlying DNN softmax on MNIST (left) and SVHN (right) in Figure~\ref{fig:out-of-distribution-confidence}. The
DkNN algorithm assigns an average  credibility of $6\%$ and $9\%$ 
to inputs from the NotMNIST and rotated MNIST test sets respectively, 
compared to $33\%$ and $31\%$ for
 the softmax probabilities.
Similar observations hold for the SVHN model: the DkNN assigns
a mean credibility of $15\%$ and $18\%$ to CIFAR-10 and rotated SVHN inputs,
in contrast with $52\%$ and $33\%$ for the softmax probabilities.

\begin{myprop}
DkNN credibility is better calibrated 
on out-of-distribution samples than 
softmax probabilities: outliers to the 
training distribution are assigned
low credibility reflecting a lack of support from  
training data.
\end{myprop}

Again, we tested here the DkNN only on ``benign'' out-of-distribution samples.
Later, we make similar observations when evaluating the DkNN on adversarial examples in Section~\ref{sec:eval-robustness}. \section{Evaluation of the Interpretability of DkNNs}
\label{sec:eval-interpretability}

The internal logic of DNNs is often controlled by a large
set of parameters, as is the case in our experiments, 
and is thus difficult to understand for a human observer.
Instead, the nearest neighbors central to 
the DkNN are an instance of \textit{explanations by example}~\cite{caruana1999case}. 
Training points whose representations are near the
test input's representation afford evidence relevant for a human
observer to rationalize the DNN's prediction.
Furthermore, research from the neuroscience community suggests
that locality-sensitivity hashing, the technique used in Section~\ref{sec:approach})
 to find nearest neighbors in the DkNN, 
may be a general principle of computation in the brain~\cite{dasgupta2017neural,valiant2014must}.

 Defining interpretability is difficult and we thus follow one of the 
evaluation methods outlined by Doshi and Kim~\cite{doshi2017towards}.
We demonstrate interpretability through a downstream practical application
of the DkNN: fairness in ML~\cite{kearns2017fair}.

Machine learning models reflect human biases, such as the ones
encoded in their training
data, which raises concerns about their lack of \textit{fairness} towards minorities~\cite{luong2011k,zafar2017fairness,zemel2013learning,kleinberg2016inherent,hardt2016equality,corbett2017algorithmic,barocas2016big}. 
This is for instance undesirable when the model's predictions are used to 
make decisions or influence humans that are taking them:
e.g., admitting a student to university~\cite{waters2014grade}, 
predicting if a defendant awaiting trial can  be released safely~\cite{kleinberg2017human},
modeling consumer credit risk~\cite{khandani2010consumer}.

We show that nearest neighbors identified by
the DkNN
help understand how training data yields model biases.
This is a step towards eliminating sources of  bias 
during DNN training.

\begin{figure}[t] 
	\centering
	\includegraphics[width=\columnwidth]{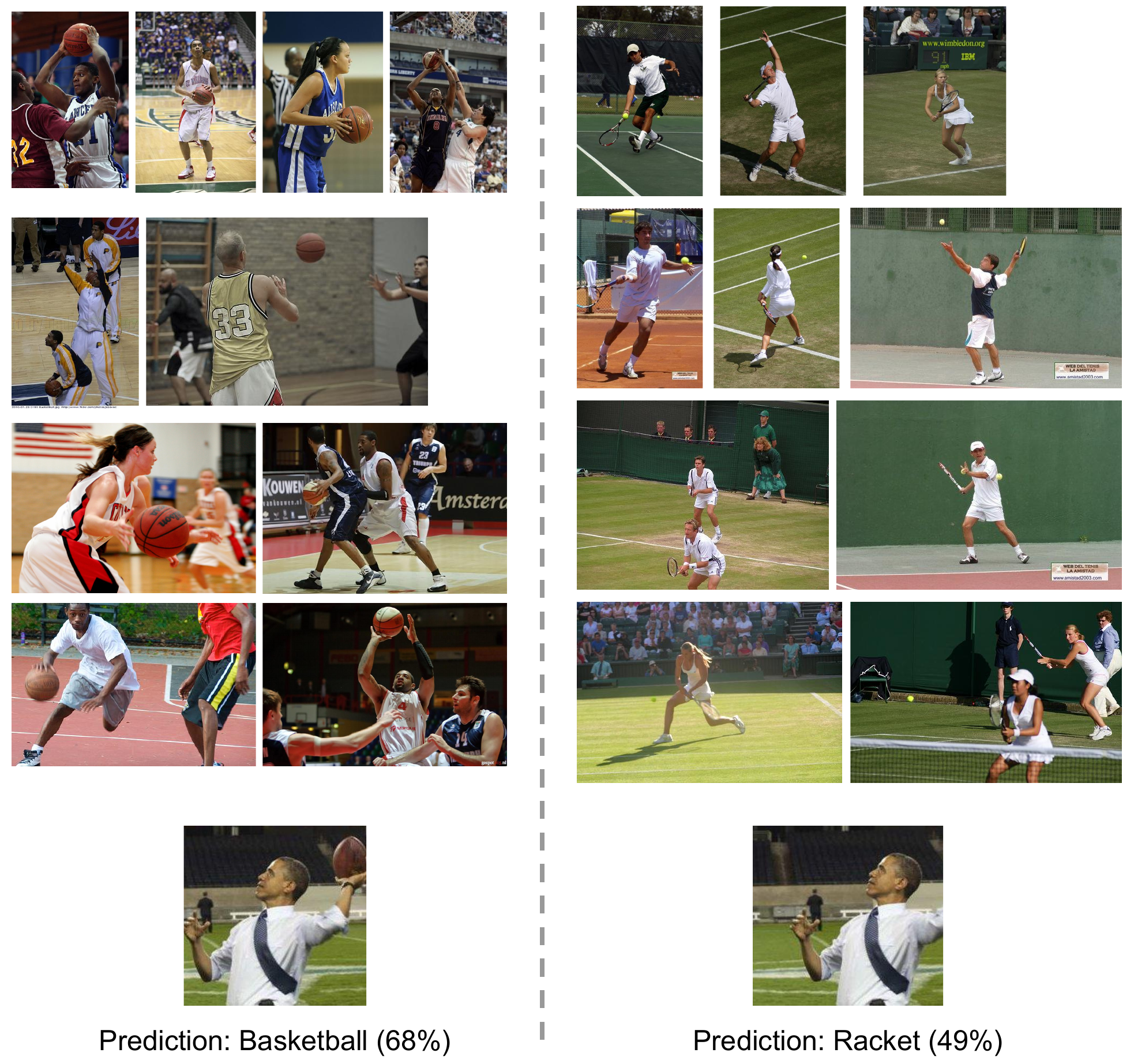} 
	\caption{\textbf{Debugging ResNet model biases}---This illustrates how the
		DkNN algorithm 
		helps to understand a bias identified by Stock and Ciss\'e~\cite{stock2017convnets}
		in the ResNet model for ImageNet. The image at the bottom of each column is the test input presented to the DkNN. Each test input is cropped slightly
		differently to include (left) or exclude (right) the football. 
		Images shown at the top are nearest neighbors in the 
		predicted class according to the representation output by the last hidden layer. This comparison suggests that the ``basketball'' prediction
		may have been a consequence of the ball being in the picture. Also note how
		the white apparel color and general arm positions of players often match
		the test image of Barack Obama.}
	\label{fig:dknn-interpretability} 
\end{figure}

Here, we consider model bias to the skin color of a person. Models for computer vision 
potentially exhibit bias towards people of
dark skin~\cite{kasperkevic2015google}. In a recent study, Stock and Ciss\'e ~\cite{stock2017convnets} demonstrate how an image
of former US president Barack Obama throwing an American football
in a stadium (see Figure~\ref{fig:dknn-interpretability})  is classified 
as a basketball 
by a popular model for computer vision---a residual neural network (ResNet~\cite{he2016identity}). 

In the following, we reproduce their experiment and apply the DkNN algorithm to this architecture
to provide explanations by example of the prediction made on
this input. To do so, we downloaded the pre-trained model from the TensorFlow repository~\cite{abadi2016tensorflow}, 
as well as the ImageNet training dataset~\cite{deng2009imagenet}.

We plot the $10$ nearest
neighbors from the training data for the class predicted by the model in
Figure~\ref{fig:dknn-interpretability}.
These neighbors are computed using the representation output by the last hidden layer
of the ResNet model. On the left, the test image that is processed by the
DNN is the same than the one used by Stock and Ciss\'e ~\cite{stock2017convnets}.
Its neighbors in the training data are  images of 7 black  and 3 white basketball players (female and male). 
Note how the basketball is similar in appearance to the football
in the image of Obama: it is of similar color and 
located in the air (often towards the top of the image).
Hence, we conjecture that the ball may play an important factor in the prediction.

We repeat the experiment with the same image cropped to remove the
football (see Figure~\ref{fig:dknn-interpretability}, right). The  prediction changes to racket. Neighbors
in this new training class are all white (female and male) tennis players.
Here again, images share several characteristics of the test image: most noticeably the
 background is always green (and a lawn) but also more subtly
the player is dressed in white and holding one of her or his hands in the air.
While this does not necessarily contradict the bias identified in prior work,
it offers alternative---perhaps complementary---explanations for the prediction made by the model.
In this particular example, in addition to the skin color of Barack Obama,
the position and appearance of the football contributed to the model's original basketball prediction.

The simplicity of the heuristic suggests that
beyond immediate benefits for human trust in deep learning,
techniques that enable interpretability---like the DkNN---will make powerful
debugging tools for deep learning practitioners to better identify the
limitations of their algorithms and models in a semi-automated way~\cite{kim2017tcav}. 
This heuristic also suggests steps towards eliminating bias in DNN training.
For instance, 
one could remove ambiguous training points or 
add new points to prevent the model
from learning an undesired correlation between a feature of the input (e.g., skin color) and one 
of the class labels (e.g., a sport).
In the interest of space, we leave a more detailed exploration of this aspect to future work.

 \section{Evaluation of the Robustness of DkNNs}
\label{sec:eval-robustness}

The lack of robustness to perturbations of their inputs is a major
 criticism faced by DNNs. Like
other ML techniques, deep learning is
for instance vulnerable to adversarial examples. Defending against these
malicious inputs is difficult because 
DNNs extrapolate with too much confidence
from their training data (as we reported in Section~\ref{ssec:out-of-distribution}). 
To illustrate, consider the example of adversarial training~\cite{szegedy2013intriguing,goodfellow2014explaining}.
The resulting models are robust to some classes
of attacks because they are trained on inputs generated by these
attacks during training but they remain vulnerable to adapted  strategies~\cite{tramer2017ensemble}.
In a direction orthogonal to defenses like adversarial training, which attempt
to have the model always output a correct prediction, we 
show here that the DkNN 
is a step towards correctly handling
malicious inputs like adversarial
examples because it:
\begin{itemize}
	\item outputs more reliable confidence estimates on adversarial examples than the softmax (Section~\ref{ssec:dknn-detect-adv-ex})
	\item provides
	insights as to why adversarial examples affect undefended
	DNNs. In the applications considered, they target the layers that automate feature extraction to introduce ambiguity that eventually builds up to significantly change the end prediction of the model despite the perturbation magnitude being small in the 
	input domain (Section~\ref{ssec:dknn-explain-adv-ex})
	\item is robust to adaptive attacks we considered, which
	modify the input to align its internal representations 
	with the ones of training points from a class that differs
	from the correct class of the input (see Section~\ref{ssec:dknn-robust-feature-adv})
\end{itemize}

\subsection{Identifying Adversarial Examples with the DkNN Algorithm}
\label{ssec:dknn-detect-adv-ex}

In Section~\ref{sec:eval-confidence}, we found that outliers to the
training distribution modeled by a DNN could
be identified at test time by ensuring that the model's internal
representations are neighbored in majority by training points whose
labels are consistent with the prediction. This is achieved by the conformal
prediction stage of the DkNN. Here, we show that
this technique is also applicable to detect
malicious inputs: e.g.,  adversarial examples. In practice, we find that
the DkNN algorithm yields well-calibrated responses on these adversarial inputs---meaning 
that the DkNN assigns low credibility to adversarial
examples unless it can recover their correct class.

\begin{table}[t]
	\centering
	{\renewcommand{\arraystretch}{1.2}
		\begin{tabular}{|c||c|c||c|c|}
			\hline
			Dataset & Attack & Attack Parameters & DNN & DkNN \\ \hline\hline
			\multirow{3}{*}{MNIST} & FGSM  & $\varepsilon\texttt{=}0.25$ & 27.1\% &  54.9\% \\\cline{2-5}
			& BIM  & $\varepsilon\texttt{=}0.25$, $\alpha\texttt{=}0.01$, $i\texttt{=}100$ & 0.7\% & 16.8\% \\\cline{2-5}
			& CW  & $\kappa\texttt{=}0$, $c\texttt{=}10^{-4}$, $i\texttt{=}2000$  & 0.7\% & 94.4\% \\\hline\hline
			\multirow{3}{*}{SVHN} & FGSM  & $\varepsilon=0.05$ & 9.3\%  & 28.6\% \\\cline{2-5}
			& BIM  & $\varepsilon\texttt{=}0.05$, $\alpha\texttt{=}0.005$, $i\texttt{=}20$ & 4.7\% & 17.9\% \\\cline{2-5}
			& CW  & $\kappa\texttt{=}0$, $c\texttt{=}10^{-4}$, $i\texttt{=}2000$ & 4.7\% & 80.5\% \\\hline\hline
			\multirow{3}{*}{GTSRB} & FGSM  & $\varepsilon=0.1$ & 12.3\%  & 22.3\% \\\cline{2-5}
			& BIM  & $\varepsilon\texttt{=}0.1$, $\alpha\texttt{=}0.005$, $i\texttt{=}20$ & 6.5\% & 13.6\% \\\cline{2-5}
			& CW  & $\kappa\texttt{=}0$, $c\texttt{=}10^{-4}$, $i\texttt{=}2000$ & 3.0\% & 74.5\% \\\hline
		\end{tabular}
		\vspace*{0.1in}
	}
	\caption{\textbf{Adversarial example classification accuracy for the DNN and DkNN}: attack parameters are chosen according to prior work. All input features were clipped to
	remain in their range. Note that most wrong predictions made by the DkNN are assigned low credibility (see Figure~\ref{fig:dknn-relia-diag-adv-ex} and the Appendix).}
	\label{tbl:adv-ex-detection}
\end{table}

\begin{figure}[t] 
	\vspace*{-0.2in}
	\begin{subfigure}[b]{0.5\columnwidth}
		\centering
		\includegraphics[width=\textwidth]{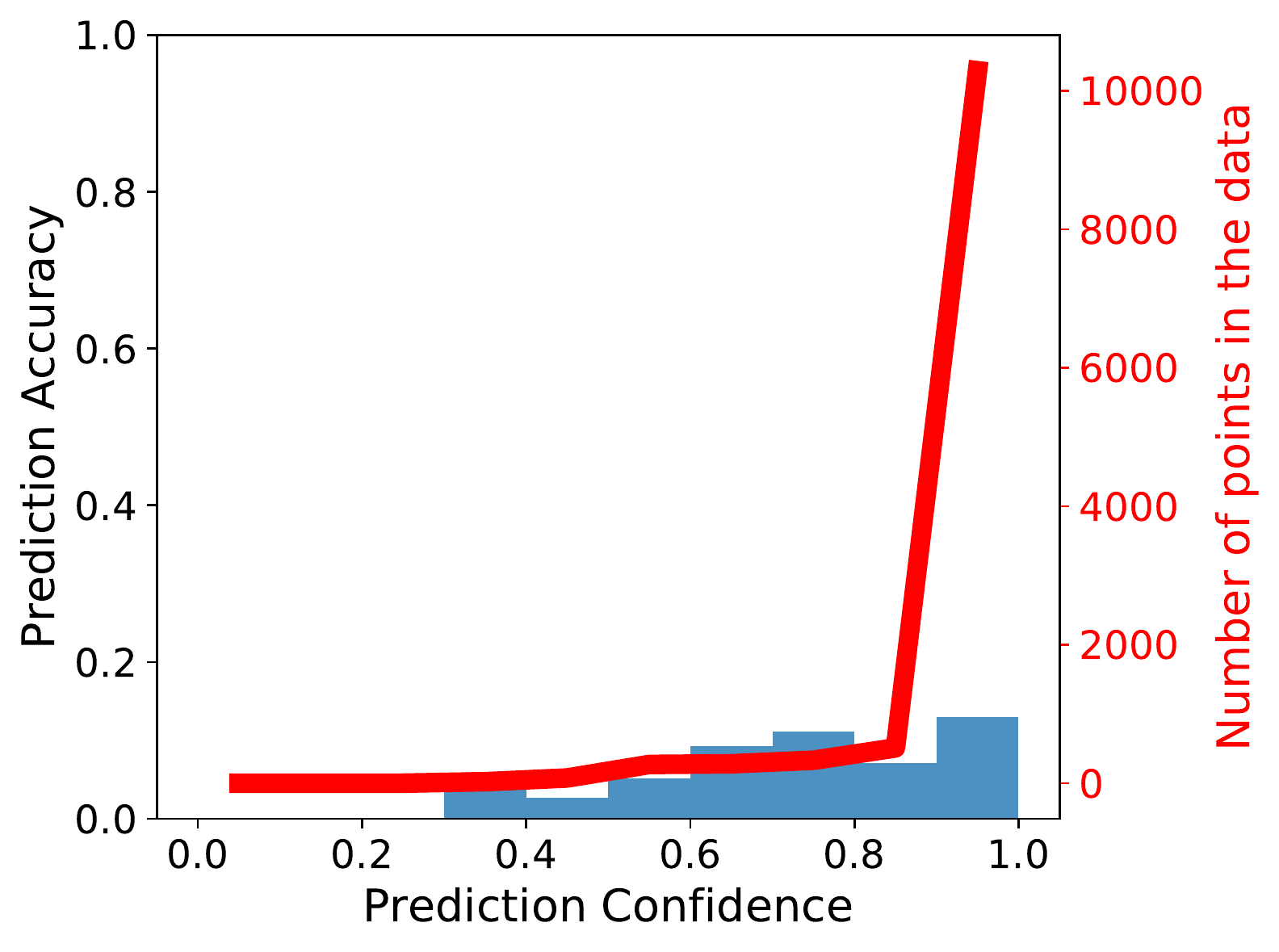} 
		\caption{Softmax - FGSM}
	\end{subfigure}~
	\begin{subfigure}[b]{0.5\columnwidth}
		\centering
		\includegraphics[width=\textwidth]{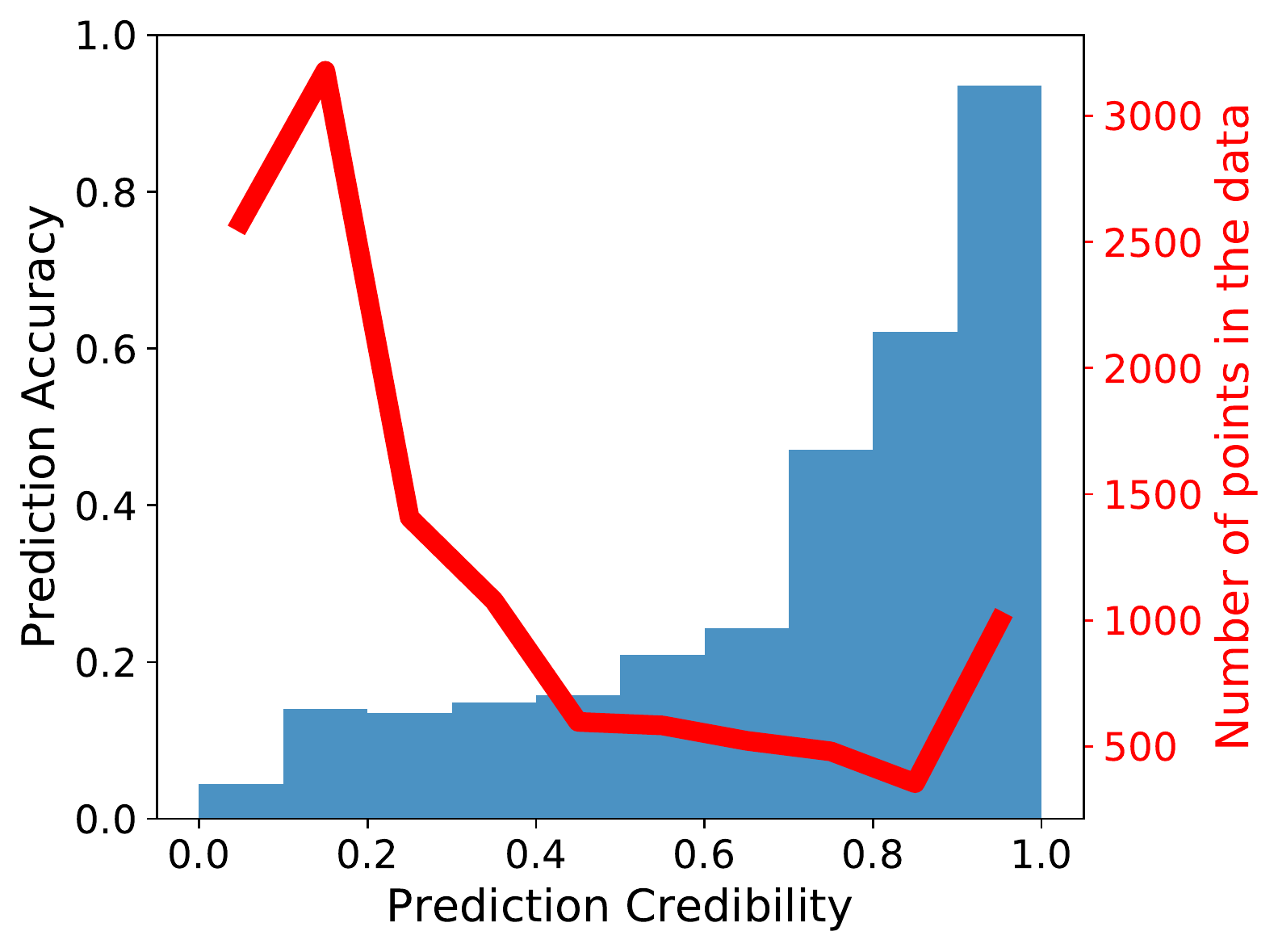} 
		\caption{DkNN - FGSM}
	\end{subfigure}
	\begin{subfigure}[b]{0.5\columnwidth}
		\centering
		\includegraphics[width=\textwidth]{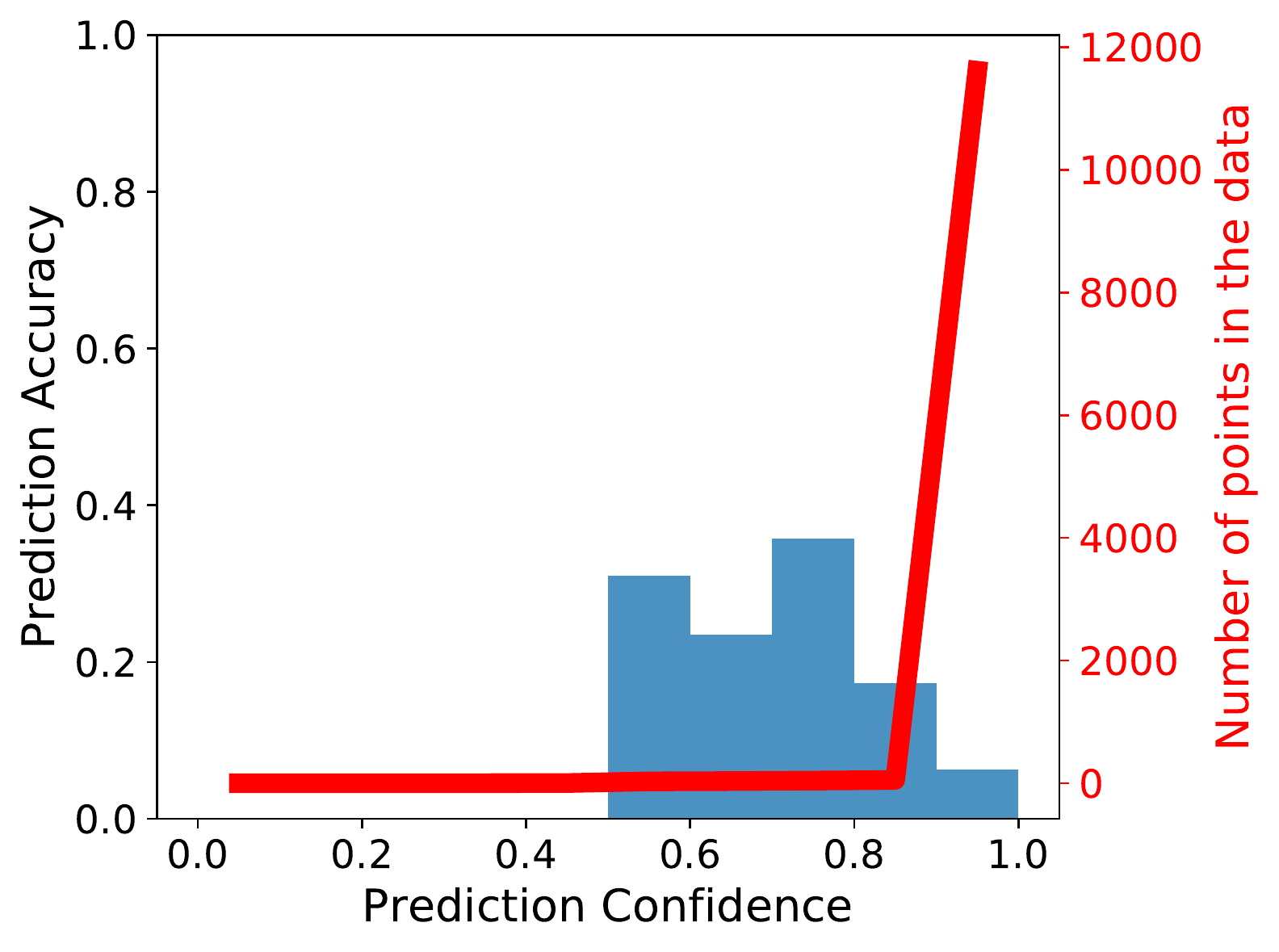} 
		\caption{Softmax - BIM}
	\end{subfigure}~
	\begin{subfigure}[b]{0.5\columnwidth}
		\centering
		\includegraphics[width=\textwidth]{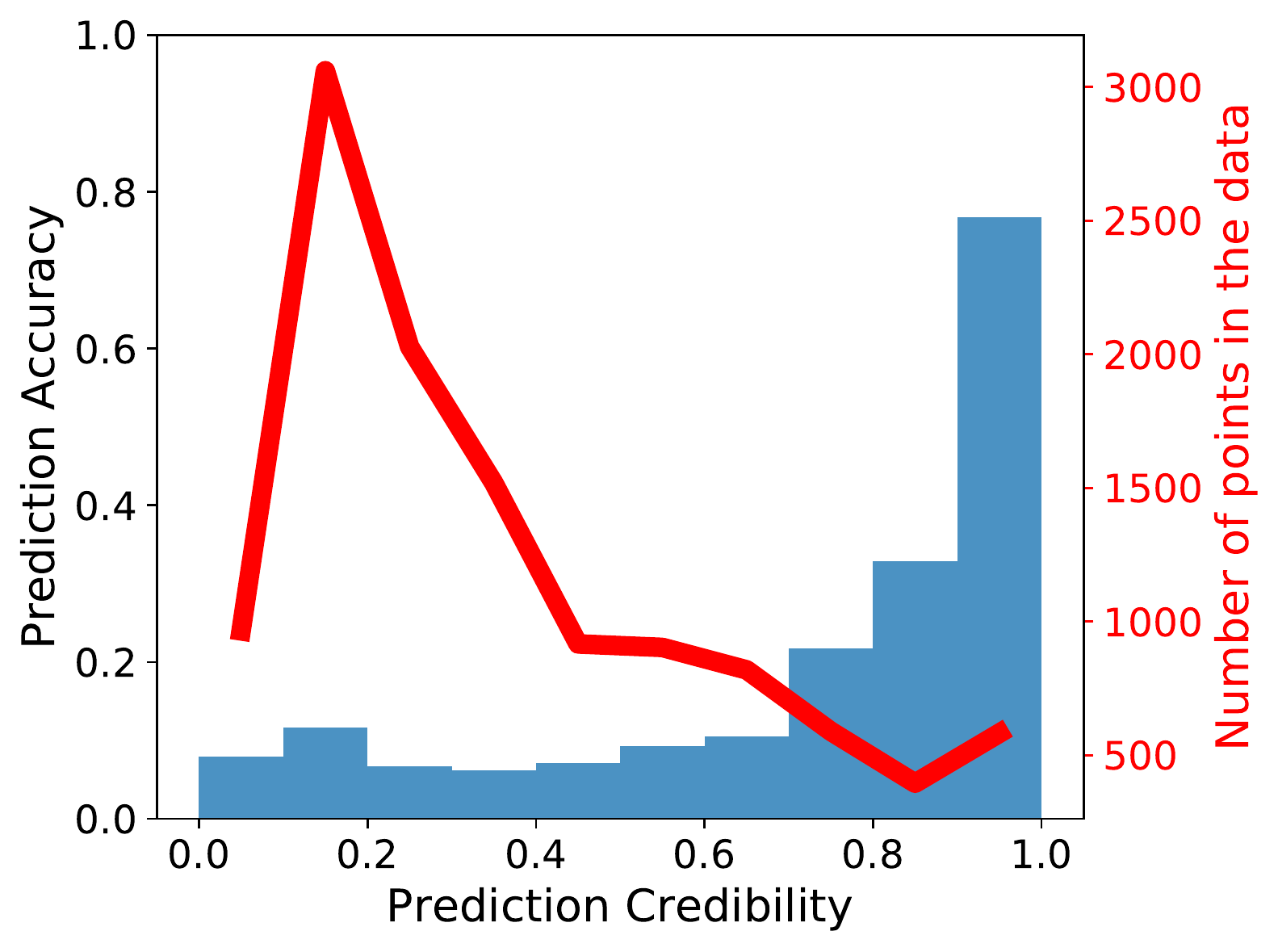} 
		\caption{DkNN - BIM}
	\end{subfigure}
	\begin{subfigure}[b]{0.5\columnwidth}
		\centering
		\includegraphics[width=\textwidth]{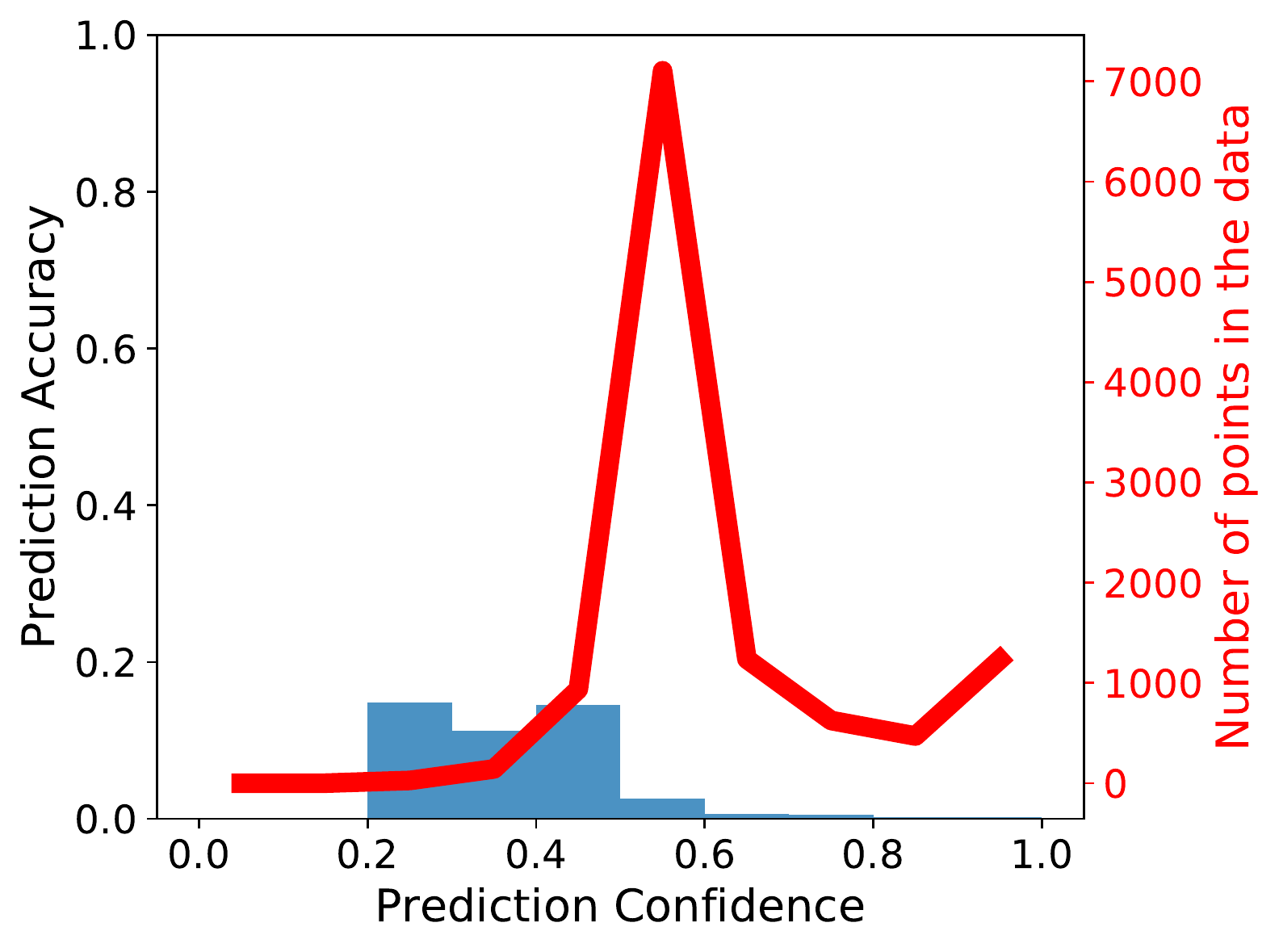} 
		\caption{Softmax - CW}
	\end{subfigure}~
	\begin{subfigure}[b]{0.5\columnwidth}
		\centering
		\includegraphics[width=\textwidth]{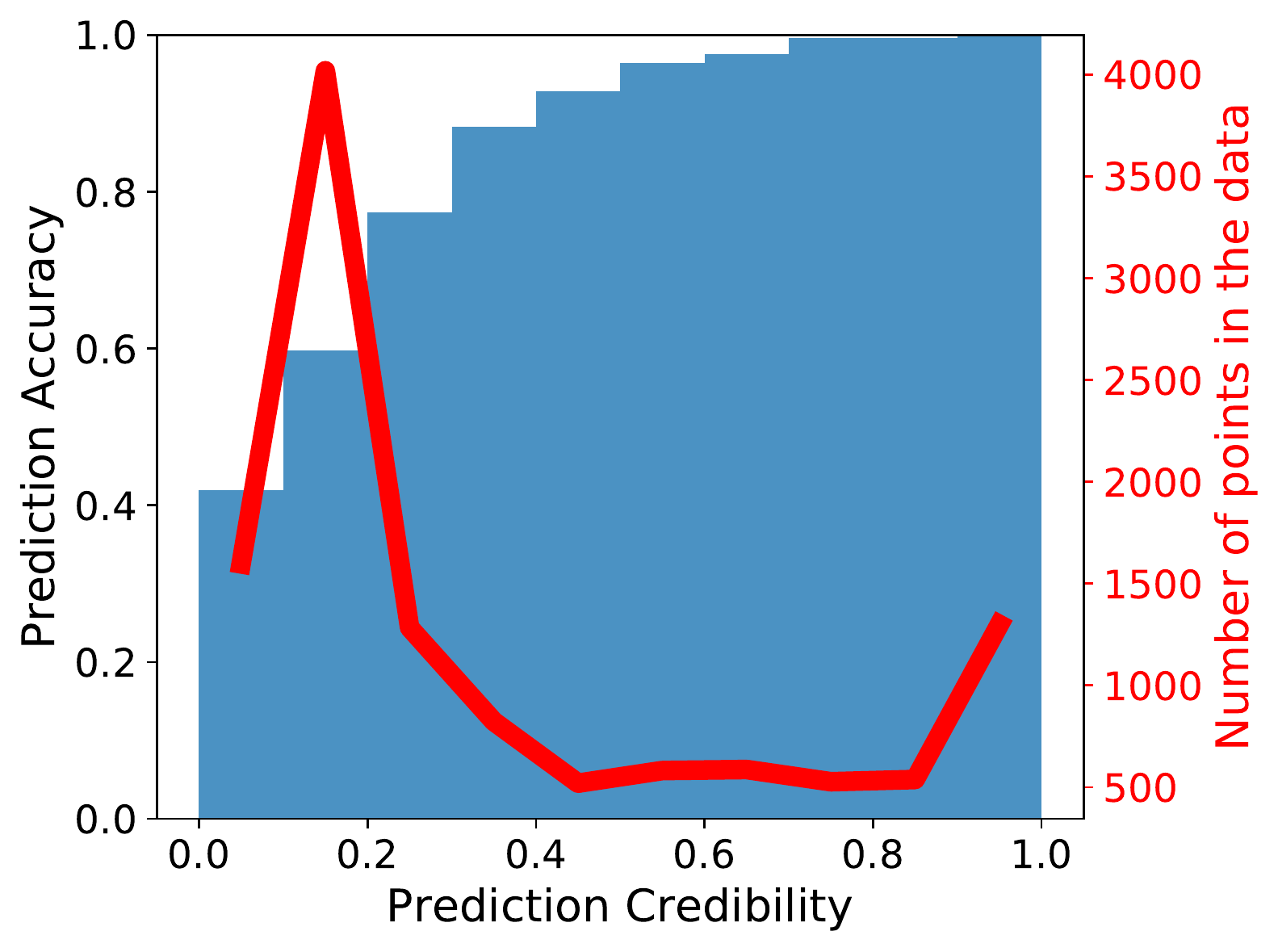} 
		\caption{DkNN - CW}
	\end{subfigure}
	\caption{\textbf{Reliability Diagrams on Adversarial Examples}---The DkNN's credibility is better calibrated (i.e., it assigns low confidence to adversarial examples) than probabilities output by the softmax of an undefended DNN. All diagrams are plotted with GTSRB test data. 
	Similar graphs for the MNIST and SVHN datasets are in the Appendix.}
	\label{fig:dknn-relia-diag-adv-ex} 
\end{figure}

We craft adversarial examples using three representative 
algorithms: the Fast Gradient Sign Method (FGSM)~\cite{goodfellow2014explaining}, 
the Basic Iterative Method (BIM)~\cite{kurakin2016adversarial},
and the Carlini-Wagner $
\ell_2$ attack (CW)~\cite{carlini2017towards}. Parameters
specific to each attack are reported in Table~\ref{tbl:adv-ex-detection}. We also include 
the accuracy of both the undefended DNN
 and the DkNN
algorithm on these inputs. 
From this, we conclude that even though attacks
successfully evade the undefended DNN, when this DNN
is integrated with the DkNN inference algorithm, some accuracy on adversarial examples
is recovered because the first layers of the DNN output
 representations on adversarial examples whose neighbors in the training
data are in the original class (the true class of the image from which adversarial
examples are crafted). 

We will come
back to this aspect in Section~\ref{ssec:dknn-explain-adv-ex} and conclude
that the ambiguity introduced by adversarial examples is marked
by a large multi-set of candidate labels in the first layers compared
to non-adversarial inputs.
However, the DkNN's error rate remains high, despite the improved performance
with respect to the underlying DNN. We now turn to the credibility of these
predictions, which we left out of consideration until now, and 
show that because the DkNN's credibility on these inputs is low, they
can largely be identified.

In Figure~\ref{fig:dknn-relia-diag-adv-ex}, we plot reliability diagrams comparing
the DkNN credibility on GTSRB adversarial examples 
with the softmax probabilities output by the DNN. 
Similar graphs for the MNIST and SVHN datasets are found in the Appendix.
Credibility is
low across all attacks for the DkNN, when compared to legitimate test points
considered in Section~\ref{sec:eval-confidence}---unless the DkNN's predicted
label is correct as indicated by the quasi-linear diagrams. 
Recall that the number of points in each bin is reflected by the red line. Hence, the DkNN outputs a credibility below 0.5 for most inputs
because predictions on
adversarial examples are not conformal with pairs of inputs and labels found in 
the training data. This behavior is a sharp departure from softmax
probabilities, which classified most adversarial examples in the wrong class
with a confidence often above 0.9 for the FGSM and BIM attacks.
We also observe that the BIM attack is
more successful than the FGSM or the CW attacks
at introducing perturbations that mislead the DkNN inference procedure. 
We hypothesize that it outputs adversarial examples that encode some
characteristics of the wrong class, which would also explain
its previously observed strong transferability
across models~\cite{kurakin2016adversarial,liu2016delving,tramer2017ensemble}.
 
\begin{myprop}
DkNN credibility
	is better calibrated than softmax probabilities. 
	When the DkNN outputs a prediction with high credibility,  this often implies
	the true label of an adversarial example was recovered.
\end{myprop}

We conclude that the good performance of credibility estimates on benign
out-of-distribution data observed in Section~\ref{sec:eval-confidence}  
is also applicable to adversarial test data considered here.  
The DkNN degrades its confidence
smoothly as adversarial examples push its inputs from legitimate points to
the underlying DNN's error region. It is even able to recover some of the
true labels of adversarial examples when the number of nearest neighboring
representations in that class is sufficiently large.

\subsection{Explaining DNN Mispredictions on Adversarial Examples}
\label{ssec:dknn-explain-adv-ex}

Nearest neighboring
representations offer insights into why DNNs are
vulnerable to small perturbations introduced by adversarial
examples. We find that adversarial examples gradually
exploit poor generalization as they are successively processed 
by each layer of the DNN: small perturbations are able
to have a large impact on the model's output because of the
non-linearities applied by each of its layers. Recall
Figure~\ref{fig:dknn-motivation}, where we illustrated how
this behavior is reflected
by neighboring representations, which gradually change from being
in the correct class---the one of the corresponding unperturbed 
test input---to the wrong class
assigned to the adversarial example.

\begin{figure}[t] 
	\centering
	\includegraphics[width=0.9\columnwidth]{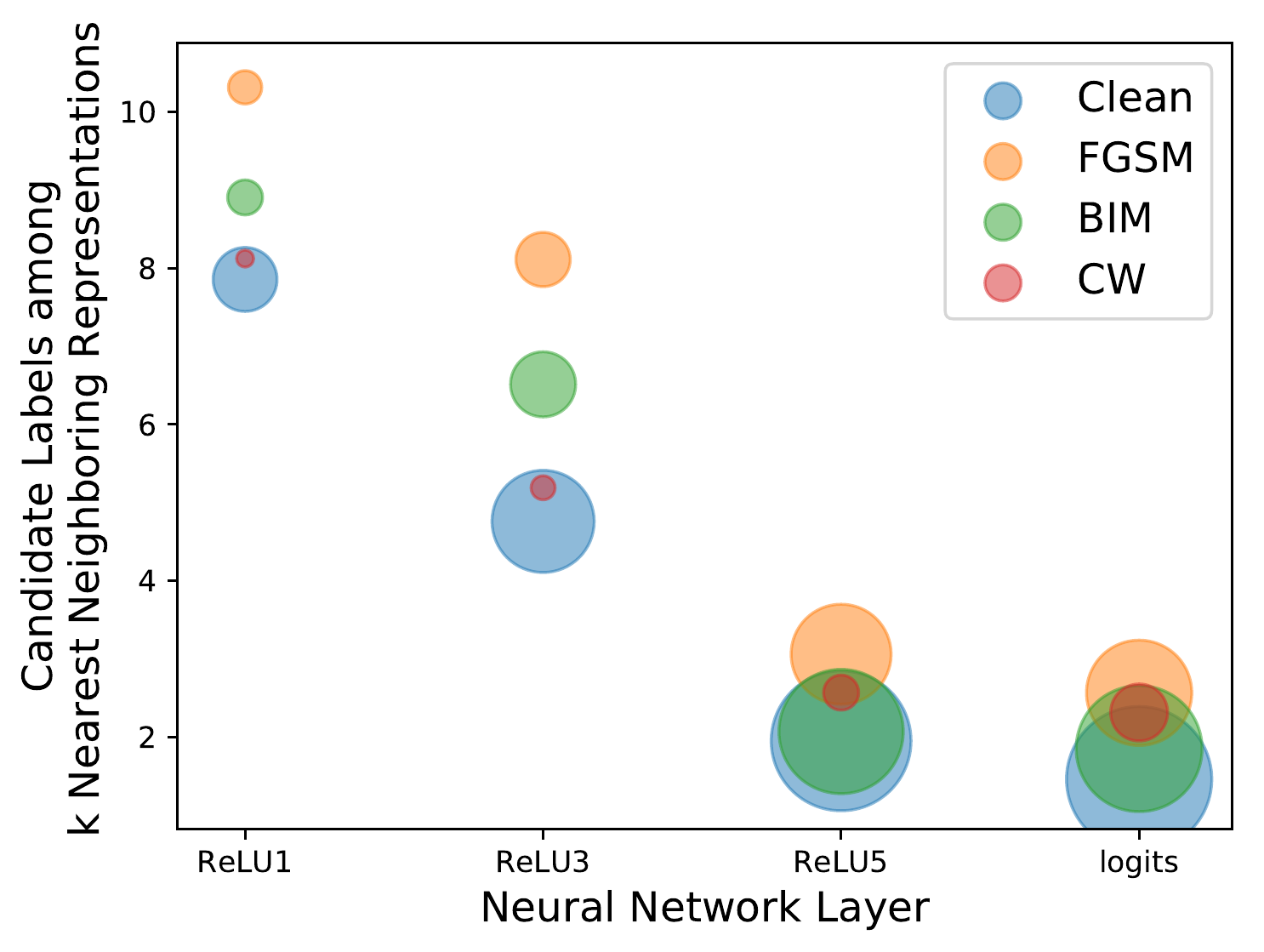}
	\caption{\textbf{Number of Candidate Labels among $k=75$ Nearest Neighboring Representations}---Shown for GTSRB with clean and adversarial data across the layers of the DNN underlying the DkNN. Points
	are centered according to the number of labels found in the neighbors; while the area of points is proportional to the number of neighbors whose label matches the DNN prediction. Representations output by lower layers of the DNN are less ambiguous for clean data than adversarial examples (nearest neighbors are more homogeneously labeled).}
	\label{fig:num-can-labels}
\end{figure}

To illustrate this, we analyze the labels
of the $k=75$ nearest neighboring training representations for
each layer when predicting on adversarial data.
We expect this number to be larger for adversarial examples than
for their legitimate counterpart because this ambiguity
would ultimately lead to the model making a mistake. 
This is
what we observe in Figure~\ref{fig:num-can-labels}. For both clean and
adversarial examples, the number of candidate labels (i.e., 
the multi-set of labels for the $k$ training points with nearest neighboring 
representations) decreases as we move up the neural architecture
from its input layer towards its output layer: the model is
projecting the input in increasingly more abstract spaces that
are better suited to classify the input. However, adversarial
examples have more candidate labels for  lower layers
than legitimate inputs: they introduce
ambiguity that is later responsible for the model's mistake.

In addition, the number of candidate labels (those of the $k=75$
nearest neighboring training representations) that match the final
prediction made by the DNN is smaller for some
attacks compared to other attacks, and for all attacks compared to legitimate inputs.
This is particularly the case for the CW attack, which is likely the
reason why the true label of adversarial examples it produces
is often recovered by the DkNN (see Table~\ref{tbl:adv-ex-detection}).
Again, this lack of conformity
between neighboring representations at different layers 
explicitly characterizes weak support for the model's 
prediction in its training data.

\begin{myprop}
Nearest neighbors provide a new avenue for measuring the strength of an attack:
if an adversarial example is able to force many of the $k$ nearest neighboring
representations to have labels in the wrong class eventually
predicted by the model, it is less likely to be detected  (e.g., by a DkNN or other techniques that may appear in the
future) and also
more likely to transfer across models.
\end{myprop}

Targeting internal representations of the DNN is the object of Section~\ref{ssec:dknn-robust-feature-adv}, where
we consider such an \textit{adaptive} attack that
targets the internal representation of the DNN.

\subsection{Robustness of the DkNN Algorithm to Adaptive Attacks}
\label{ssec:dknn-robust-feature-adv}

Our experimental results suggest that we should
not only study the vulnerability of DNNs as
a whole, but also at the level of their hidden layers. This is the goal of \textit{feature adversaries} introduced by Sabour et 
al.~\cite{sabour2015adversarial}. 
Rather than forcing a model
to misclassify, these adversaries produce adversarial examples
that force a DNN to output an internal representation
that is close to the one it outputs on a \textit{guide input}.
Conceptually, this attack may be deployed for any of the hidden layers
that make up modern DNNs. For instance, 
if the adversary is interested in attacking layer $l$, it
solves the following optimization problem:
\vspace*{-0.08in}
\begin{equation}
\label{eq:feature-adversaries}
x^* = \arg\min_{x} \| f_l(x) - f_l(x^*) \| \text{ s.t. } \|x-x^*\| \leq \varepsilon
\vspace*{-0.08in}
\end{equation}
where the norm is typically chosen to be the $\ell_\infty$ norm.

This strategy is a natural
 candidate for an adaptive attack against our
DkNN classification algorithm. An adversary aware that the
defender is using the DkNN algorithm to strengthen the
robustness of its predictive model needs to
produce adversarial examples that are not only (a) misclassified by
the DNN that underlies the DkNN algorithm but also (b) closely
aligned with internal representations of the training data for the
class that the DNN is mistakenly classifying the adversarial example in. 

Hence, we evaluate 
the DkNN against feature adversaries.
We assume a strong, perhaps insider, adversary with knowledge of the
training set used by the defender. This  ensures that we consider
the worst-case setting for deploying our DkNN algorithm
and not rely on security by obscurity~\cite{kerckhoffs1883cryptographic}.

Specifically, given a test point $x$, we target the first
layer $l=1$ analyzed by the DkNN, e.g., the output
of the first convolutional layer in our experiments on 
MNIST and SVHN.
In our feature adversaries attack, we let the guide input 
be the input from a different class whose representation
at layer $l=1$
is closest from the input we are attempting to attack.
This heuristic returns a guide input that
is already semantically close to the input being attacked,
making it easier to find smaller pertubations in the
input domain (here, the pixel domain) that forces the predicted
representation of the input being attacked to match
the representation of the guide input.
We then run the attack proposed by Sabour et al.~\cite{sabour2015adversarial} to find
an adversarial input according to this guide and n
test the prediction of our DkNN when it
is presented with the 
adversarial input. 

\begin{figure}[t] 
	\centering
	\includegraphics[width=0.79\columnwidth]{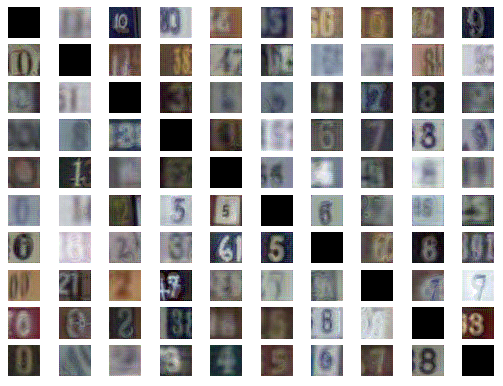} 
	\caption{\textbf{Feature Adversarial Examples against our DkNN algorithm}---Shown for SVHN (see Appendix for MNIST). Adversarial examples are organized according to their original label (rows) and the DkNN's prediction (columns).}
	\label{fig:feature-adv-ex}
\end{figure} 

Figure~\ref{fig:feature-adv-ex} shows a set of adversarial
images selected according to their order of appearance
SVHN test set (a similar figure for MNIST is found in the Appendix). Images are laid out 
on the grid such that the \textit{rows} indicate the class they were
originally from (that is the correct class for the input that
was attacked) and \textit{columns} correspond to the model's prediction
on the adversarial image. 
Although the adversarial images depicted evade our DkNN algorithm
in the sense that the images are not classified in the class of the
original input they were computed from, the perception of
a human is also affected significantly: all images are either ambiguous or modified so much
that the wrong (predicted) class is now drawn in the
image. In other words, when the attack succeeded 
on MNIST (19.6\% of the inputs at $\varepsilon=0.3$) and SVHN 
(70.0\% of the inputs at $\varepsilon=0.1$), it
altered some semantics in order to
have the adversarial input's representation match the representation of the guide input
from a different class. This can be explained by the fact that
the adversary needs to target the representation output by the first
layer in order to ensure the DkNN algorithm will find nearest neighbors
in the predicted class when analyzing this layer.

The ambiguity of many adversarial images returned by the
feature adversarial attack---together with the small
perturbation in $\ell_\infty$ norm that these images have ($0.3$
for MNIST and $0.1$ for SVHN)---raises some
questions about the methodology commonly followed in the
literature to evaluate adversarial example attacks and 
defenses. Indeed, the robustness of a machine learning model
for a computer vision application (e.g., image classification)
is often defined as its ability to constantly predict the
same class for all inputs found within an $\ell_p$ norm ball centered
in any of the test set's inputs. The existence of inputs
that are ambiguous to the human's visual system in this
ball suggests that 
we should establish a different definition
of the adversarial space that characterizes human
perception more accurately (perhaps one of the metrics used to
evaluate compression algorithms for instance~\cite{wang2004image}).
Note that this is not the case in other application domains,
such as adversarial examples for malware detection~\cite{laskov2014practical,grosse2016adversarial},
where an algorithmic oracle is often available---such as
a virtual machine running the executable as shown
in Xu et al.~\cite{xu2016automatically}. 
This suggests that future work should propose new definitions
that not only characterize robustness with
respect to inputs at test time but also in terms of the
model's training data.

 \section{Conclusions}
\label{sec:conclusions}

We introduced the Deep k-Nearest Neighbors (DkNN) algorithm, which
inspects the internals of a deep neural network (DNN) at test time
to provide \textit{confidence}, \textit{interpretability} and \textit{robustness} properties.
The DkNN algorithm compares  layer representation predictions with the nearest neighbors used to train the model.
The resulting credibility measure assesses conformance of representation prediction with the training data.  When
the training data and prediction are in agreement, the prediction is likely to be accurate.  If the  prediction and training data are not in agreement, then the prediction does not have the training data support to be credible. This is the case with inputs that are ambiguous (e.g., some inputs contain multiple classes or are partly occluded due to imperfect preprocessing) or were maliciously perturbed by an adversary to produce an adversarial example. Hence, this characterization of confidence that spans the hierarchy of representations within of a DNN ensures the integrity of the model.
The neighbors also enable
interpretability of model predictions because they are points
in the input domain that serve as support for the prediction and are easily understood and  interpreted by
human observers. 

Our findings highlight the benefits of integrating simple
inference procedures as ancillary validation of the predictions of complex learning algorithms.  Such validation is a potentially new avenue  
to provide security in machine learning systems. We anticipate that
many open problems at the intersection of machine learning
and security will benefit from this perspective, including
\textit{availability} and \textit{integrity}.  We are excited to  explore these and other related areas in the near future.

 \section*{Acknowledgments}

The authors thank \'Ulfar Erlingsson for essential discussions 
on internal DNN representations and visualization of the DkNN.
We also thank Ilya Mironov for very helpful discussions on the nearest
neighbors in high dimensional spaces and detailed comments on a
draft. 
The authors are grateful for comments
by Mart\'in Abadi, Ian Goodfellow, Harini Kannan, Alex Kurakin and Florian Tram\`er on a draft.
We also thank Megan McDaniel for taking good care of our diet in the last
stages of writing.

Nicolas Papernot is supported
by a Google PhD Fellowship in Security. Some of the GPU equipment used
in our experiments was donated by NVIDIA.
Research was supported by the Army Research Laboratory,
under Cooperative Agreement Number W911NF-13-2-0045 (ARL Cyber Security
CRA), and the Army Research Office under grant W911NF-13-1-0421.
The views and conclusions contained in this document are those of the
authors and should not be interpreted as representing the official policies,
either expressed or implied, of the Army Research Laboratory or the U.S.
Government. The U.S.\ Government is authorized to reproduce and distribute
reprints for government purposes notwithstanding any copyright notation hereon. 

\bibliographystyle{IEEEtran}

\newpage

\appendix

\subsection{Model architectures}

Models were trained
with Adam at a learning rate of $10^{-3}$.

\begin{table}[h]
	\centering
	{\renewcommand{\arraystretch}{1.2}
		\begin{tabular}{|c|c||c|c|c|}
			\hline
			Layer & Layer Parameters  & MNIST & SVHN & GTSRB \\ \hline\hline
			Conv 1 & 64 filters, (8x8), (2x2), same & Y & Y & Y \\ \hline
			Conv 2 & 128 filters, (6x6), (2x2), valid & Y & Y & Y \\ \hline
			Conv 3 & 128 filters, (5x5), (1x1), valid & Y & Y & Y \\ \hline
			Linear & 200 units & N & N & Y \\ \hline
			Linear & 10 units & Y & Y & Y \\ \hline
		\end{tabular}
		\vspace*{0.1in}
	}
	\caption{\textbf{DNN architectures for evaluation}: the last three
	columns indicate which layers are used for the different datasets. All  architectures were implemented using TensorFlow~\cite{abadi2016tensorflow} and CleverHans~\cite{papernot2016cleverhans}. Parameters for convolutions are in the
	following order: filters, kernel shape, strides, and padding.}
	\label{tbl:model-architectures}
\end{table}

\subsection{Reliability diagrams}

We provide here additional reliability diagrams for the MNIST and GTSRB datasets.
The experimental setup used to generate them is described in
Section~\ref{ssec:dknn-detect-adv-ex}, along with an interpretation of the diagrams.

\begin{figure}[h] 
		\centering
	\begin{subfigure}[b]{0.5\columnwidth}
			\centering
		\includegraphics[width=0.8\textwidth]{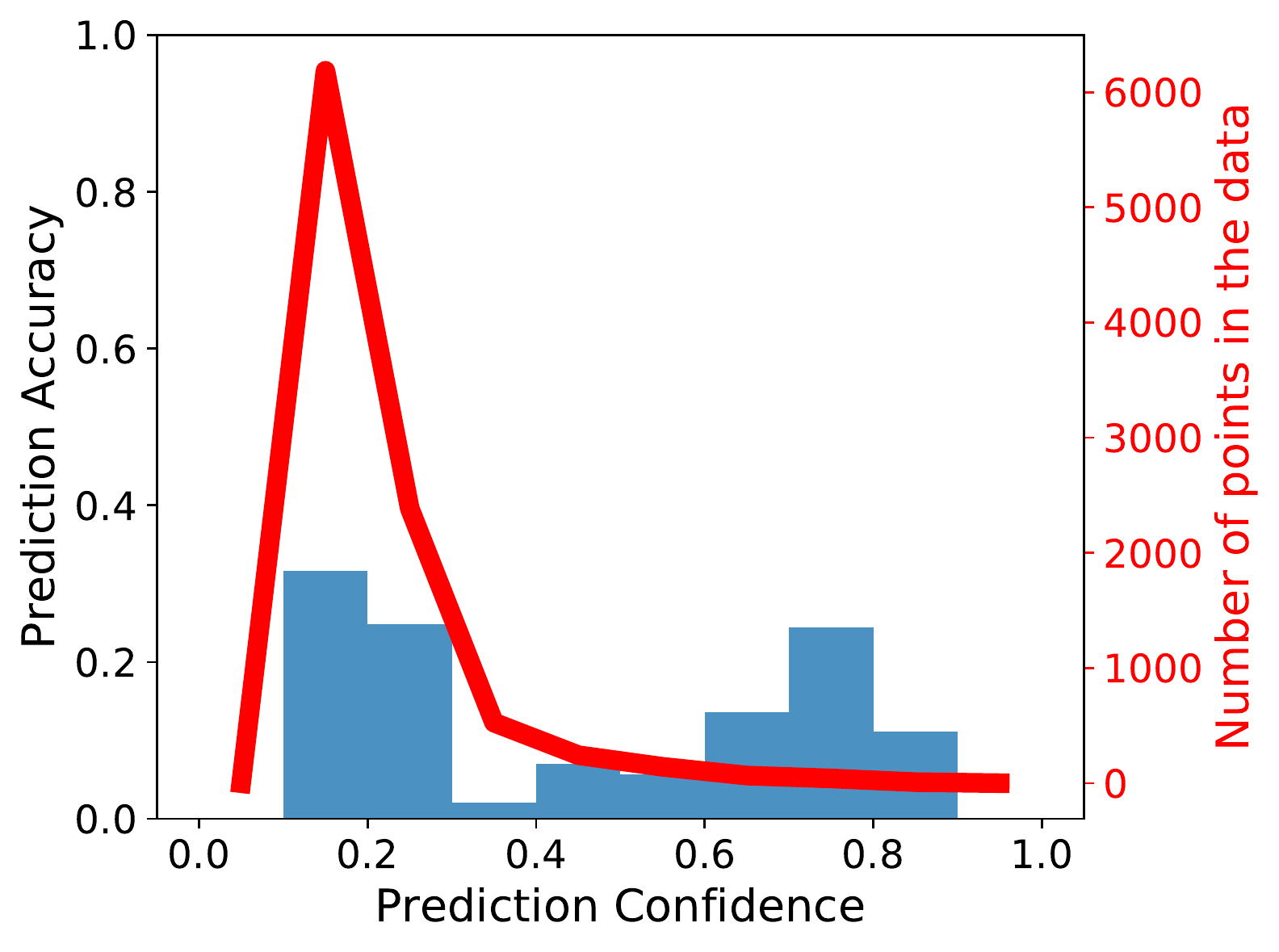} 
		\caption{Softmax - FGSM}
	\end{subfigure}~
	\begin{subfigure}[b]{0.5\columnwidth}
		\centering
		\includegraphics[width=0.8\textwidth]{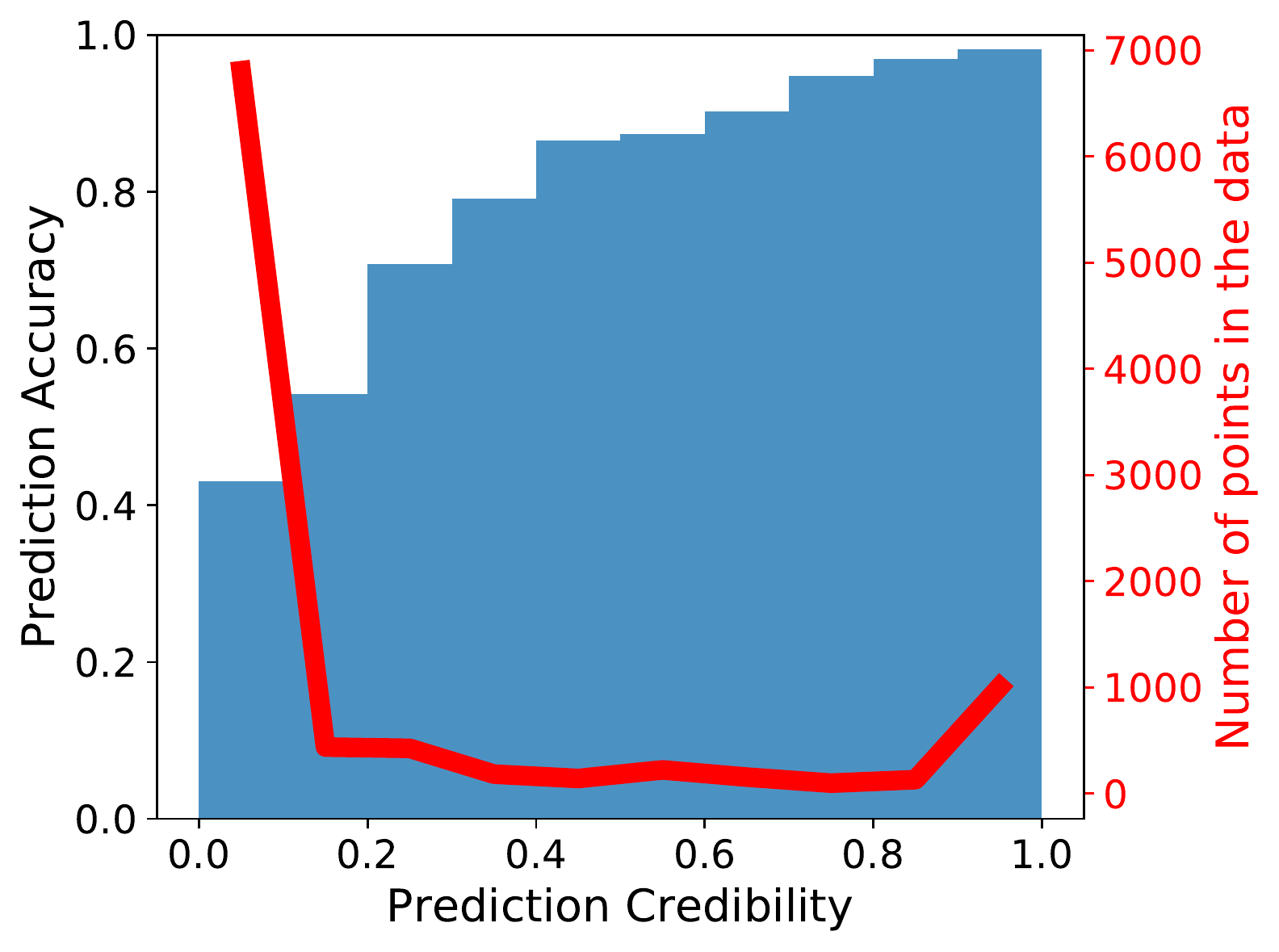} 
		\caption{DkNN - FGSM}
	\end{subfigure}
	\begin{subfigure}[b]{0.5\columnwidth}
		\centering
		\includegraphics[width=0.8\textwidth]{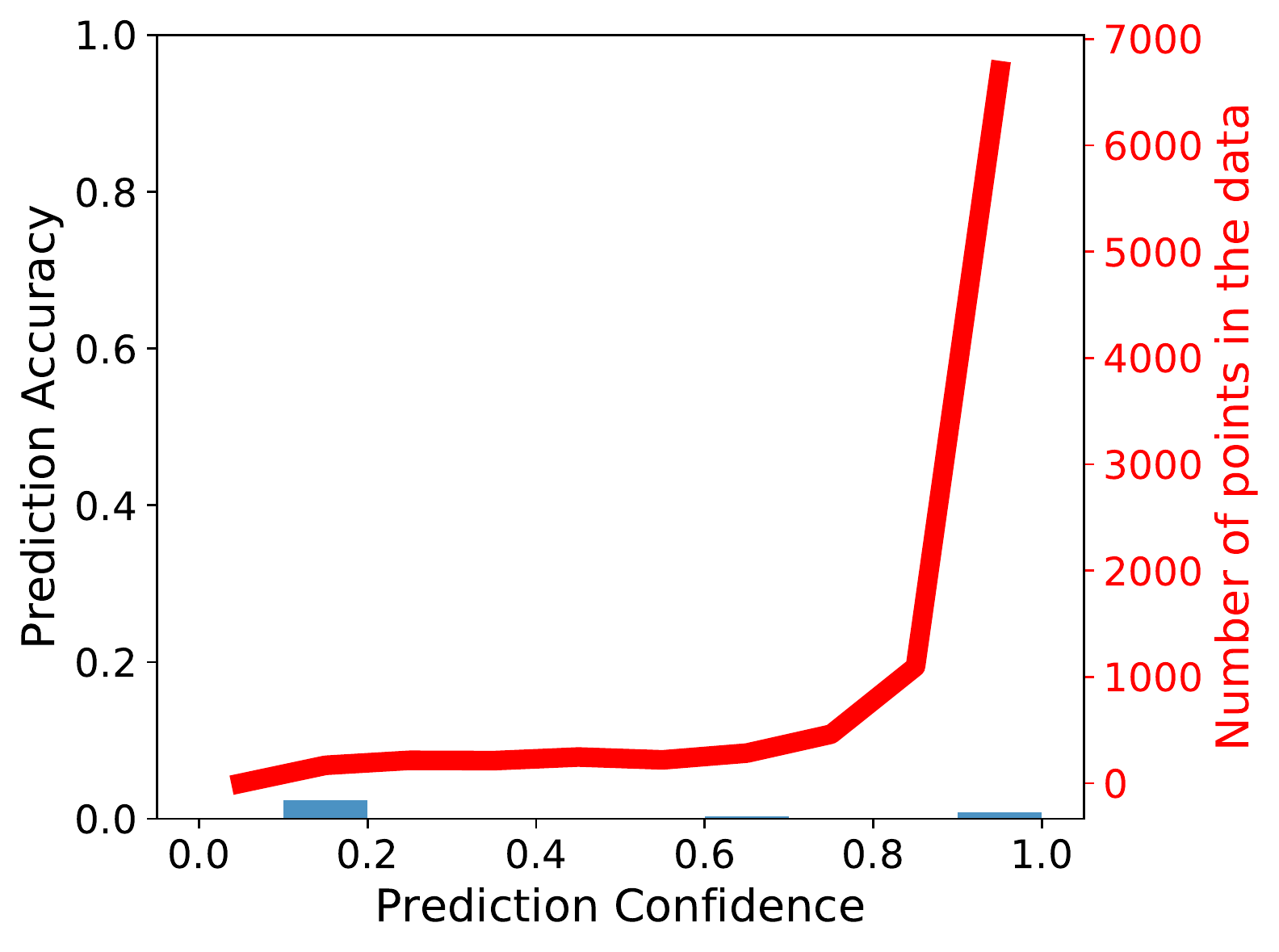} 
		\caption{Softmax - BIM}
	\end{subfigure}~
	\begin{subfigure}[b]{0.5\columnwidth}
		\centering
		\includegraphics[width=0.8\textwidth]{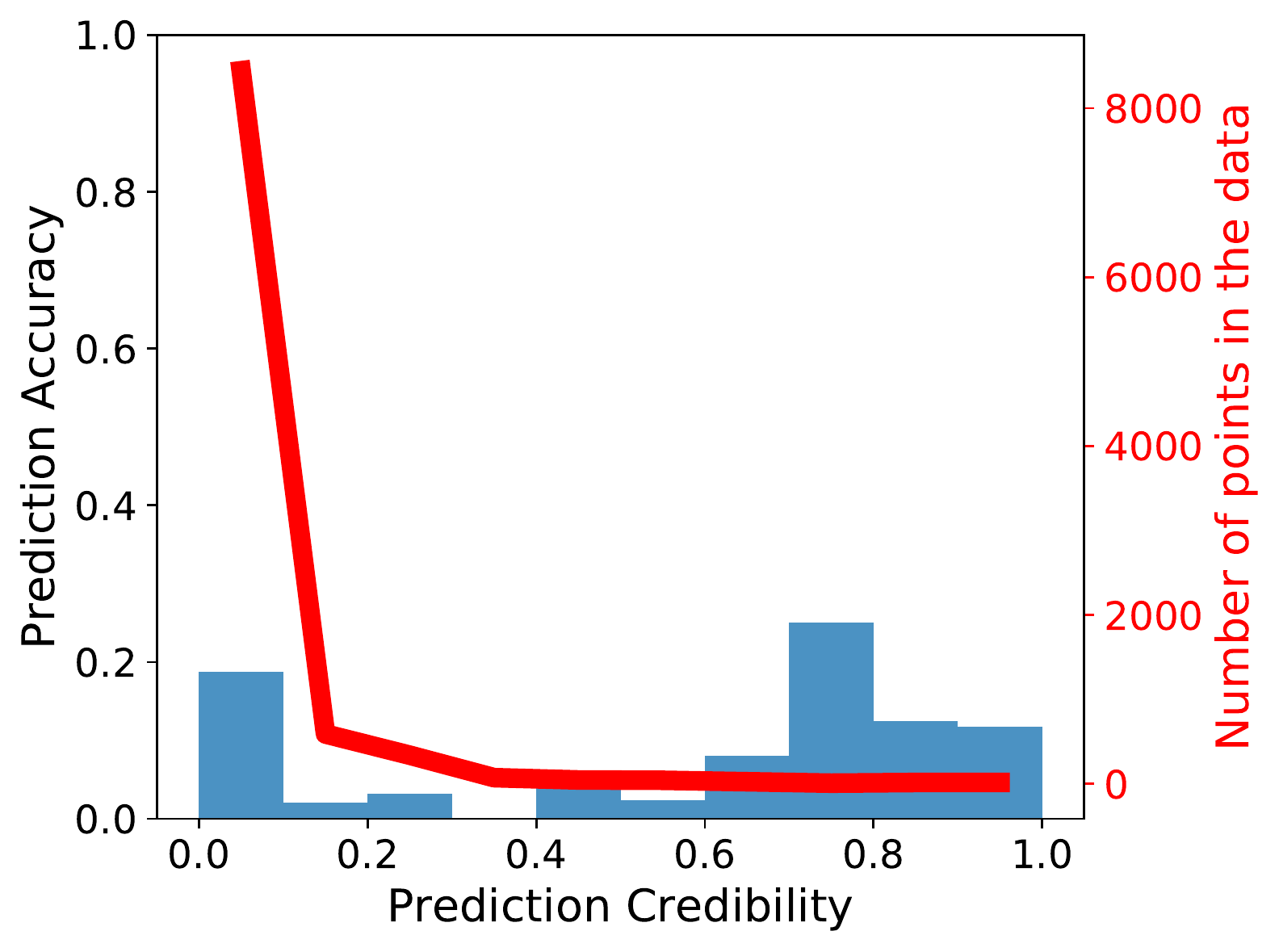} 
		\caption{DkNN - BIM}
	\end{subfigure}
	\begin{subfigure}[b]{0.5\columnwidth}
		\centering
		\includegraphics[width=0.8\textwidth]{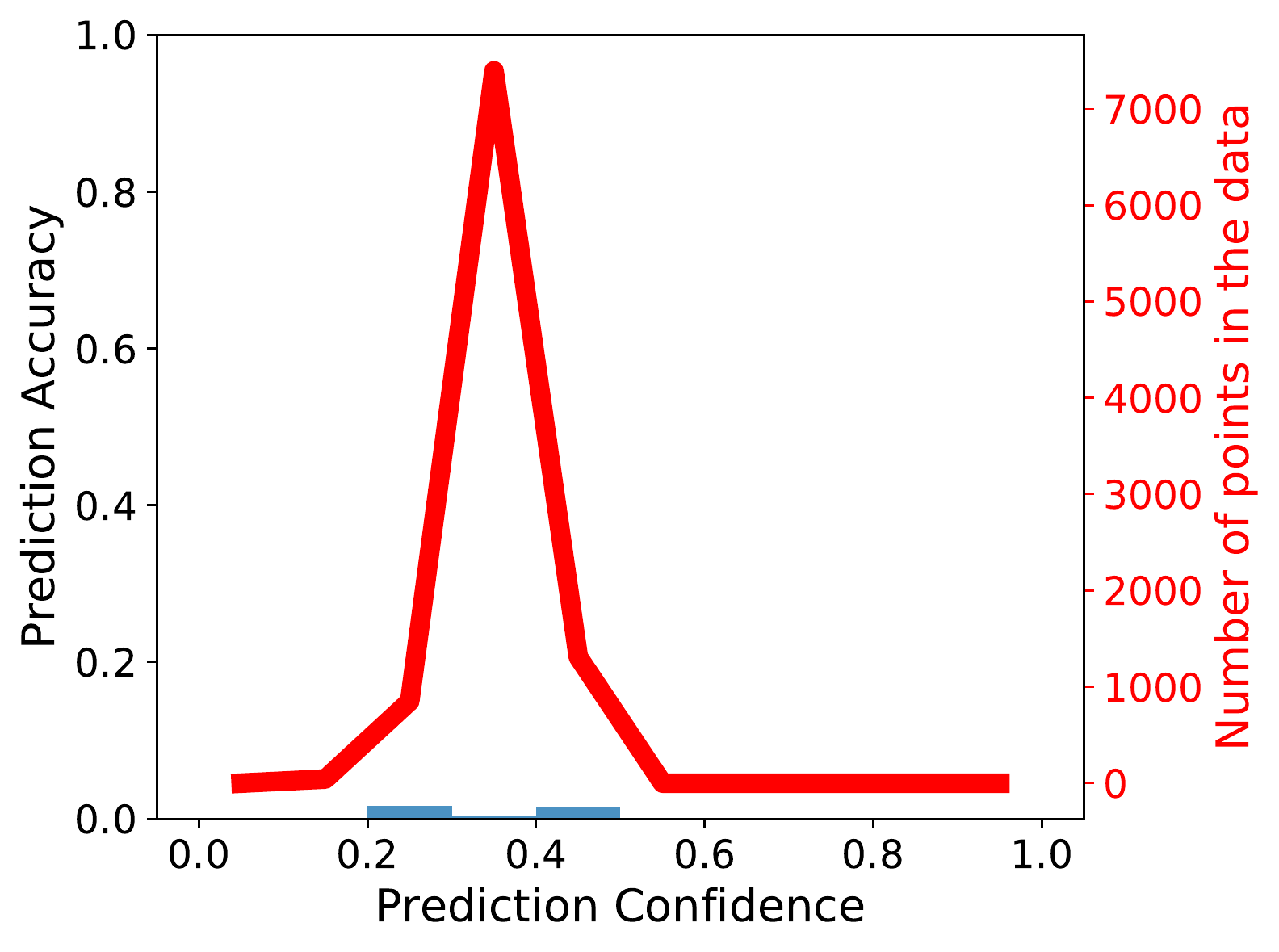} 
		\caption{Softmax - CW}
	\end{subfigure}~
	\begin{subfigure}[b]{0.5\columnwidth}
		\centering
		\includegraphics[width=0.8\textwidth]{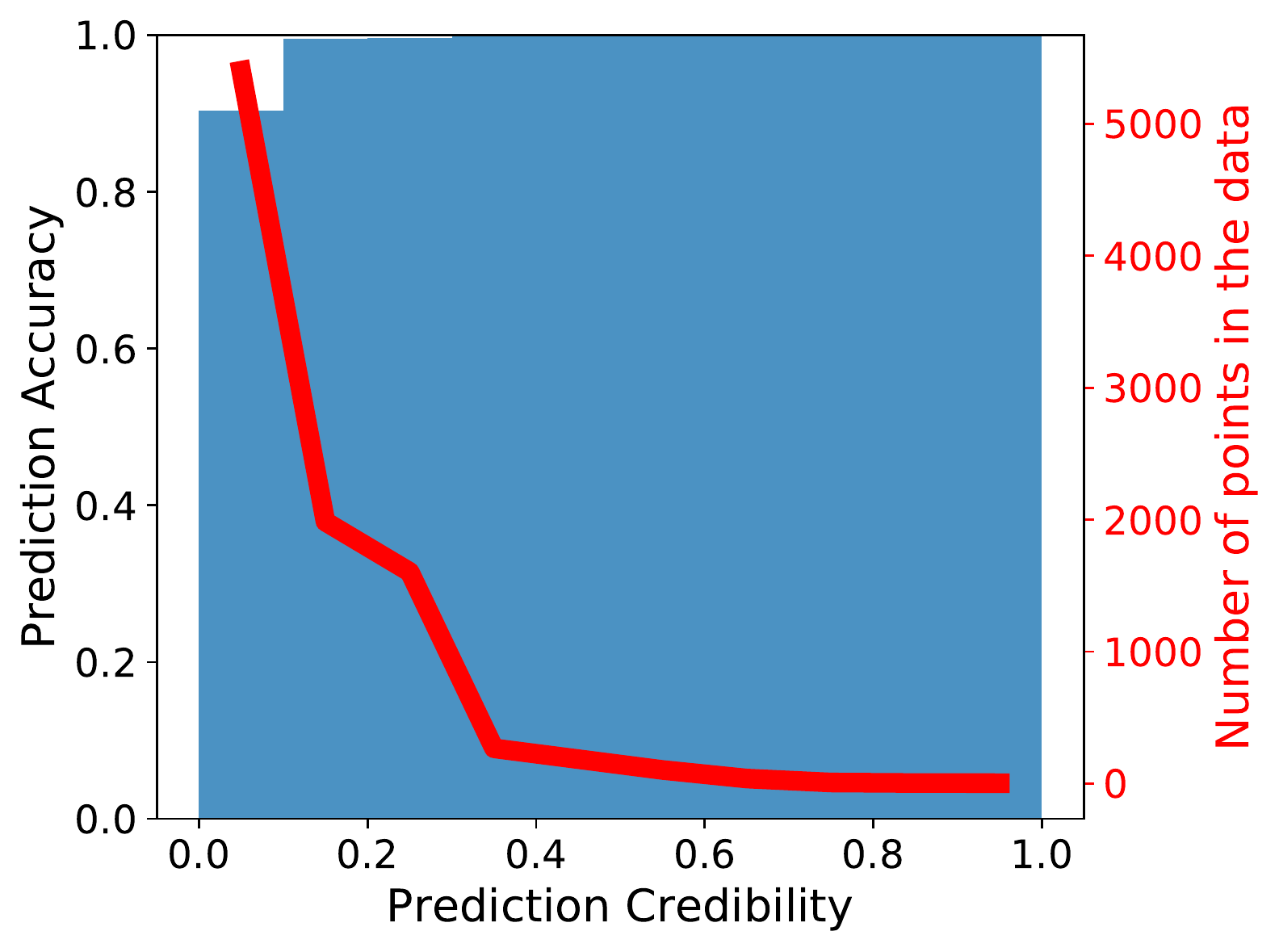} 
		\caption{DkNN - CW}
	\end{subfigure}
	\caption{\textbf{Reliability Diagrams on Adversarial Examples}---All diagrams are for MNIST test data, see Figure~\ref{fig:dknn-relia-diag-adv-ex}  for details.}
\end{figure}

\begin{figure}[h] 
	\begin{subfigure}[b]{0.5\columnwidth}
		\centering
		\includegraphics[width=0.8\textwidth]{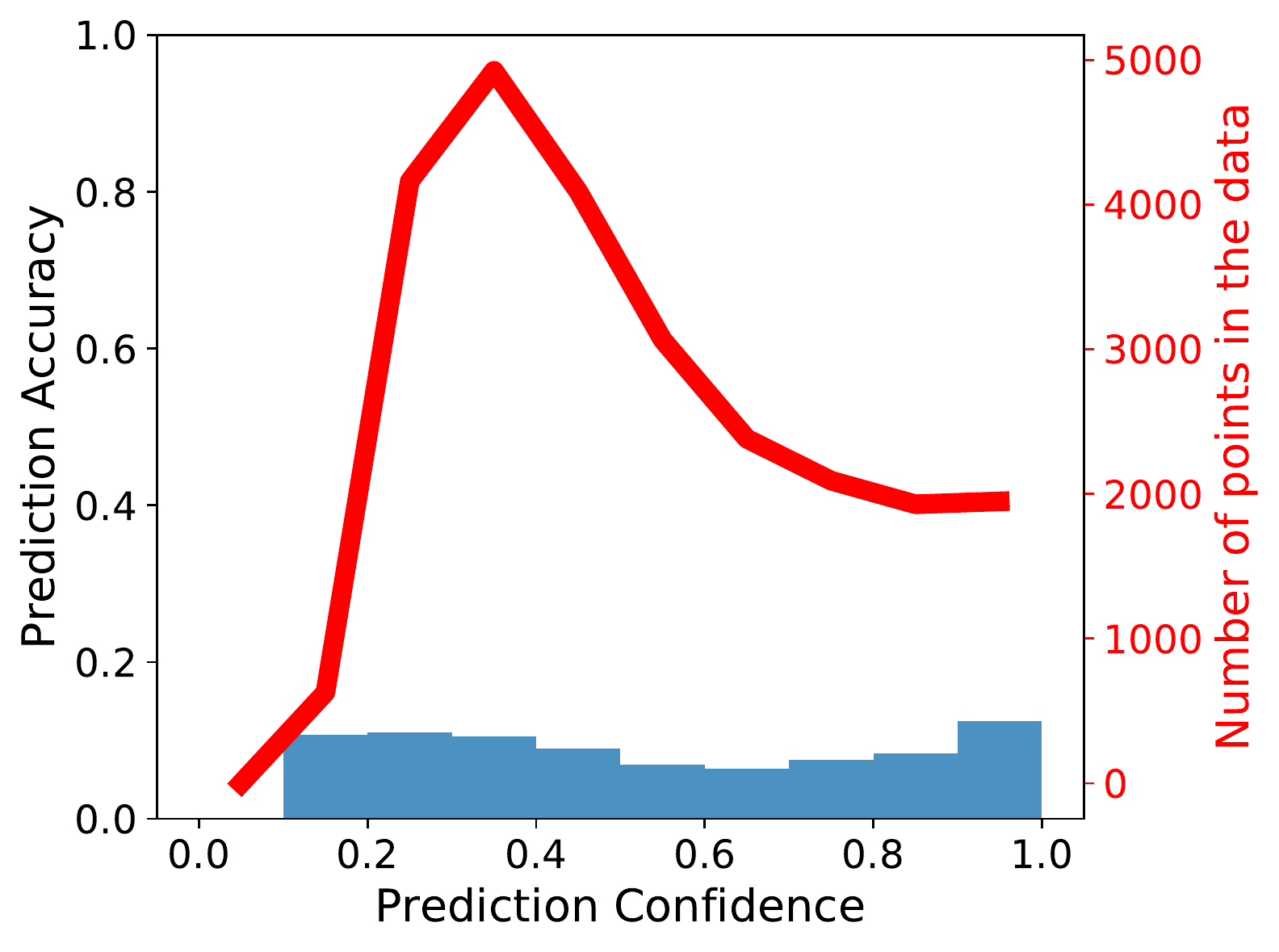} 
		\caption{Softmax - FGSM}
	\end{subfigure}~
	\begin{subfigure}[b]{0.5\columnwidth}
		\centering
		\includegraphics[width=0.8\textwidth]{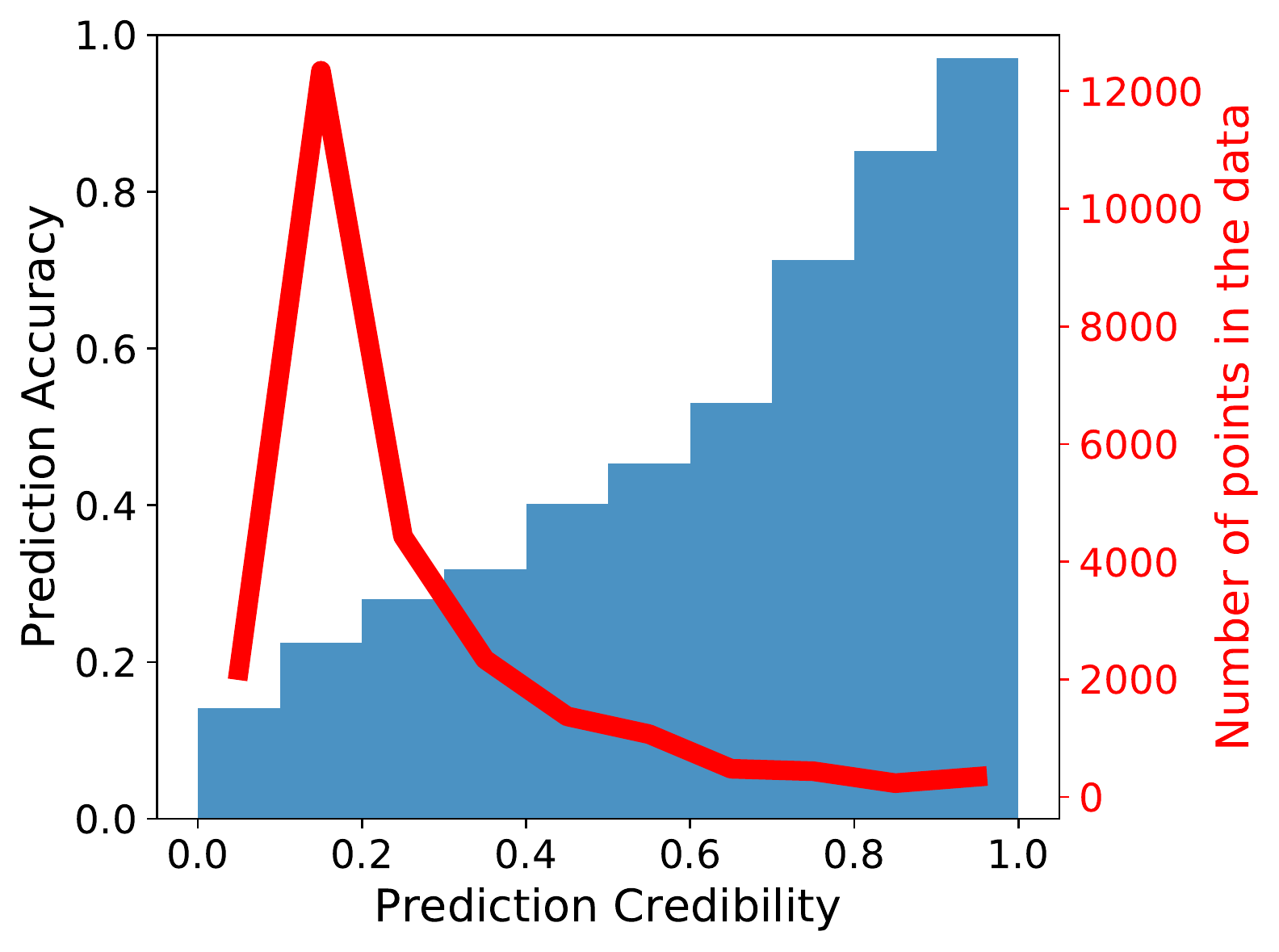} 
		\caption{DkNN - FGSM}
	\end{subfigure}
	\begin{subfigure}[b]{0.5\columnwidth}
		\centering
		\includegraphics[width=0.8\textwidth]{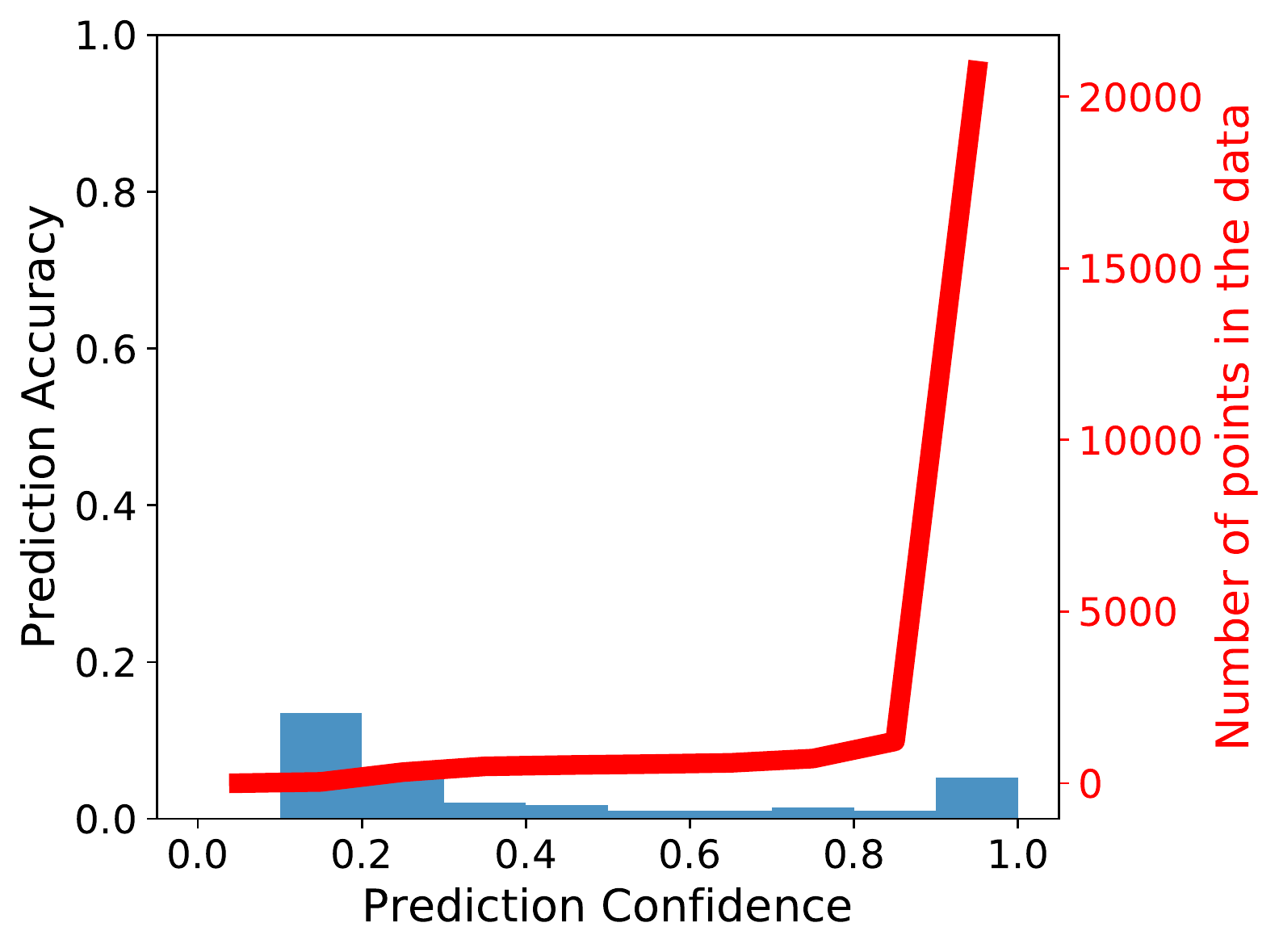} 
		\caption{Softmax - BIM}
	\end{subfigure}~
	\begin{subfigure}[b]{0.5\columnwidth}
		\centering
		\includegraphics[width=0.8\textwidth]{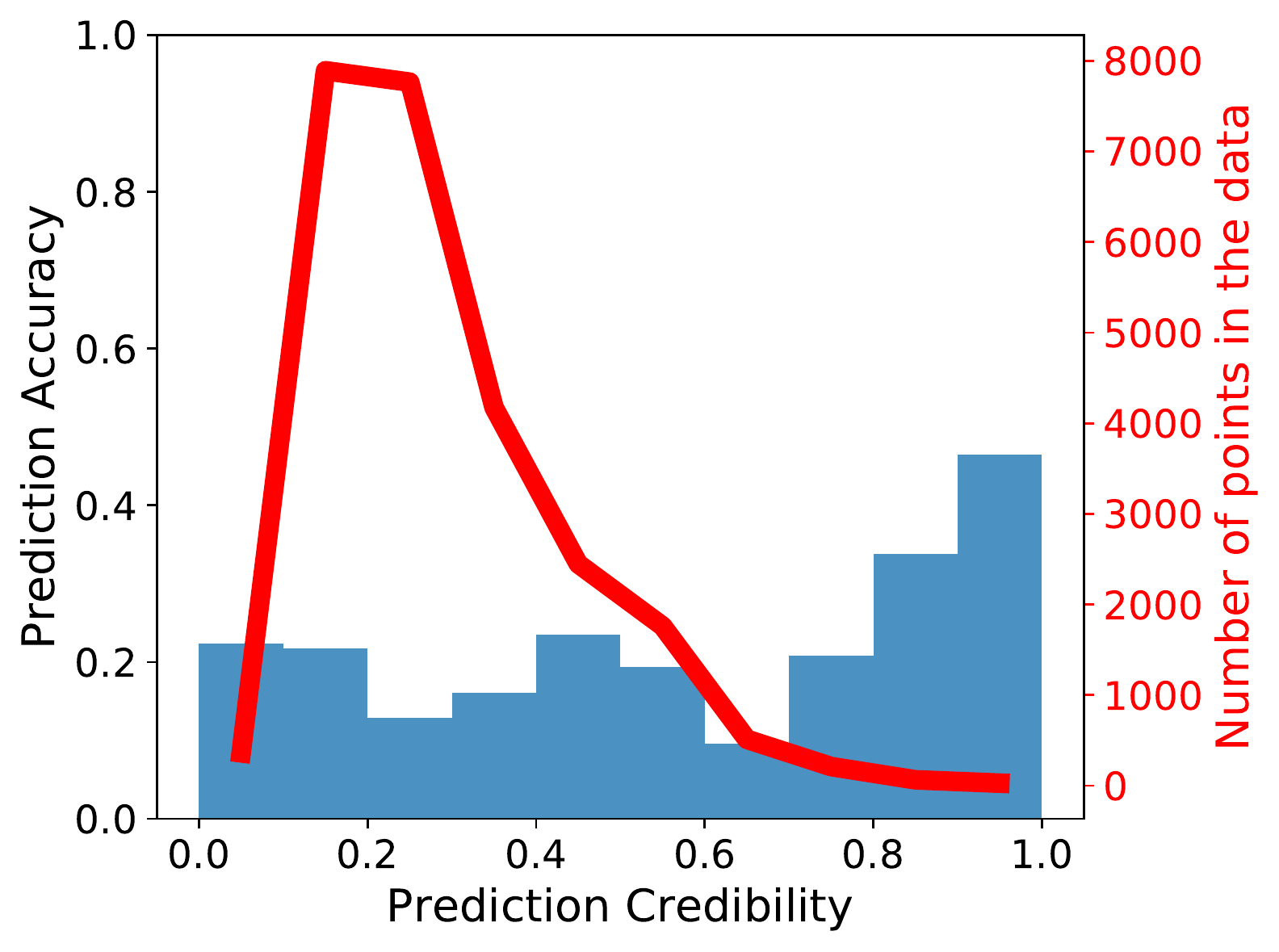} 
		\caption{DkNN - BIM}
	\end{subfigure}
	\begin{subfigure}[b]{0.5\columnwidth}
		\centering
		\includegraphics[width=0.8\textwidth]{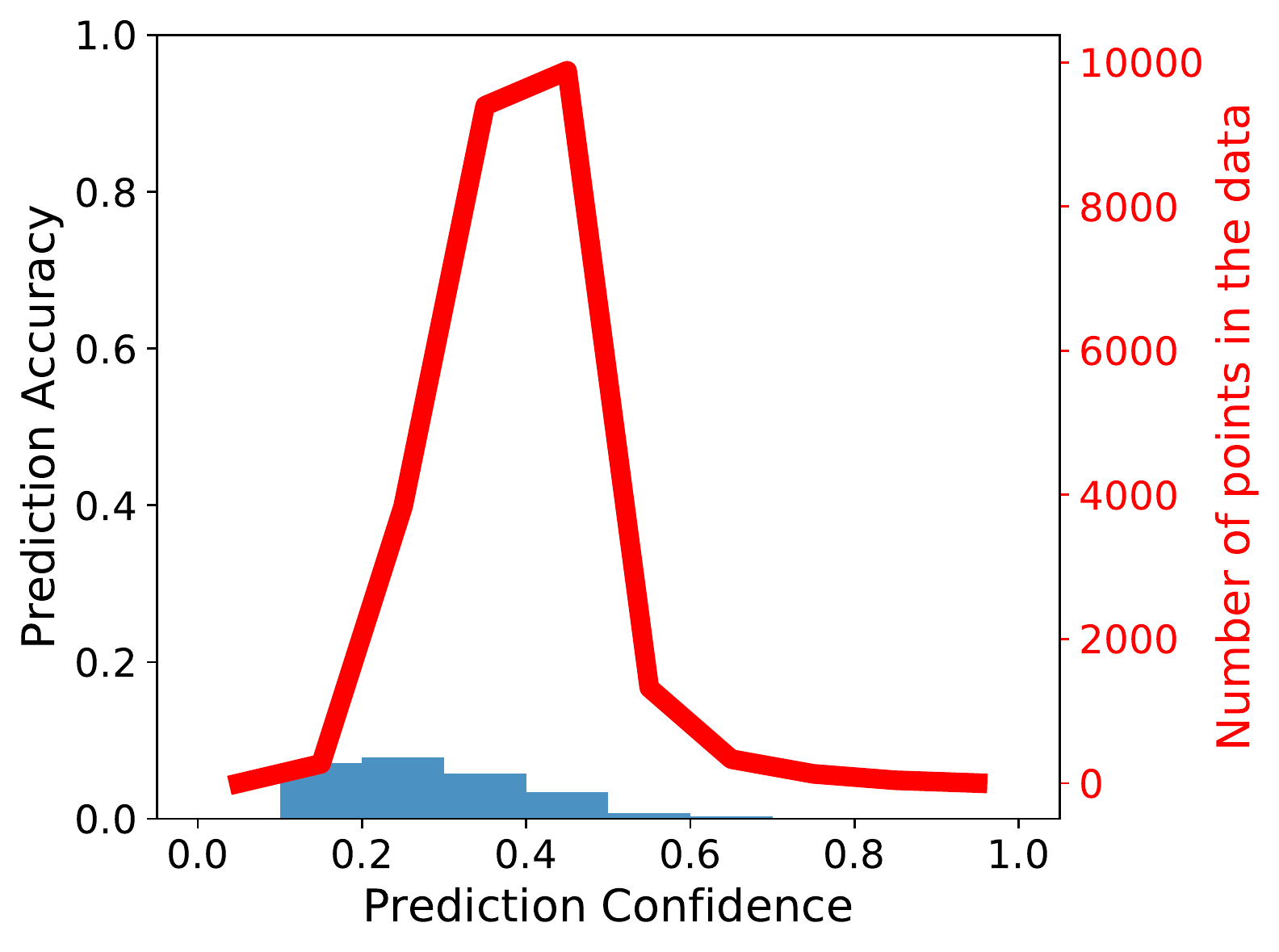} 
		\caption{Softmax - CW}
	\end{subfigure}~
	\begin{subfigure}[b]{0.5\columnwidth}
		\centering
		\includegraphics[width=0.8\textwidth]{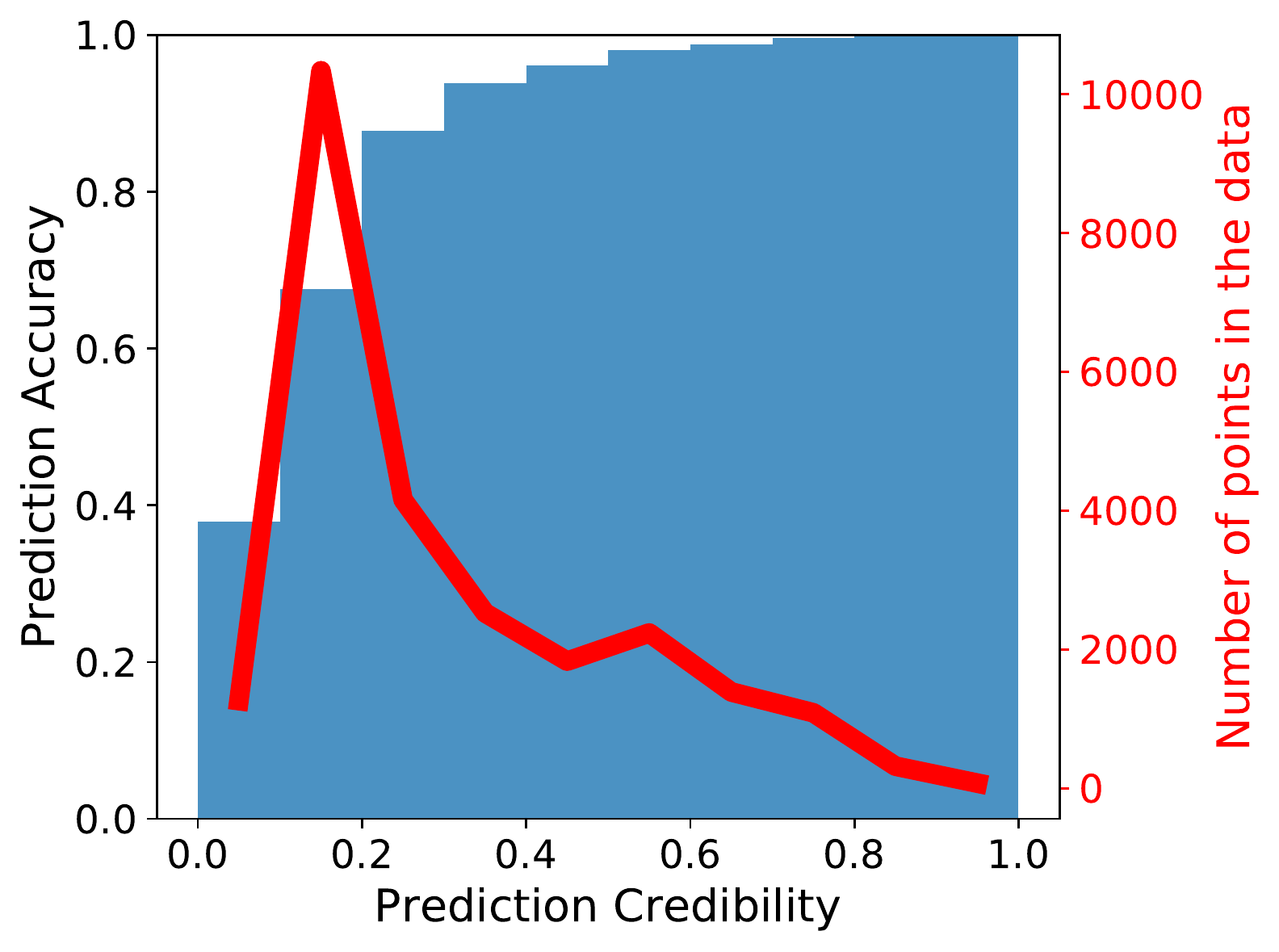} 
		\caption{DkNN - CW}
	\end{subfigure}
	\caption{\textbf{Reliability Diagrams on Adversarial Examples}---All diagrams are for SVHN test data, see Figure~\ref{fig:dknn-relia-diag-adv-ex}  for details.}
\end{figure}

\newpage
\subsection{Analysis of the nearest neighbors}

We provide here additional diagrams analyzing the labels of nearest neighbors
found on the clean and adversarial data for the MNIST and SVHN datasets.
The experimental setup used to generate them is described in
Section~\ref{ssec:dknn-explain-adv-ex}. 

\begin{figure}[h] 
	\centering
	\includegraphics[width=\columnwidth]{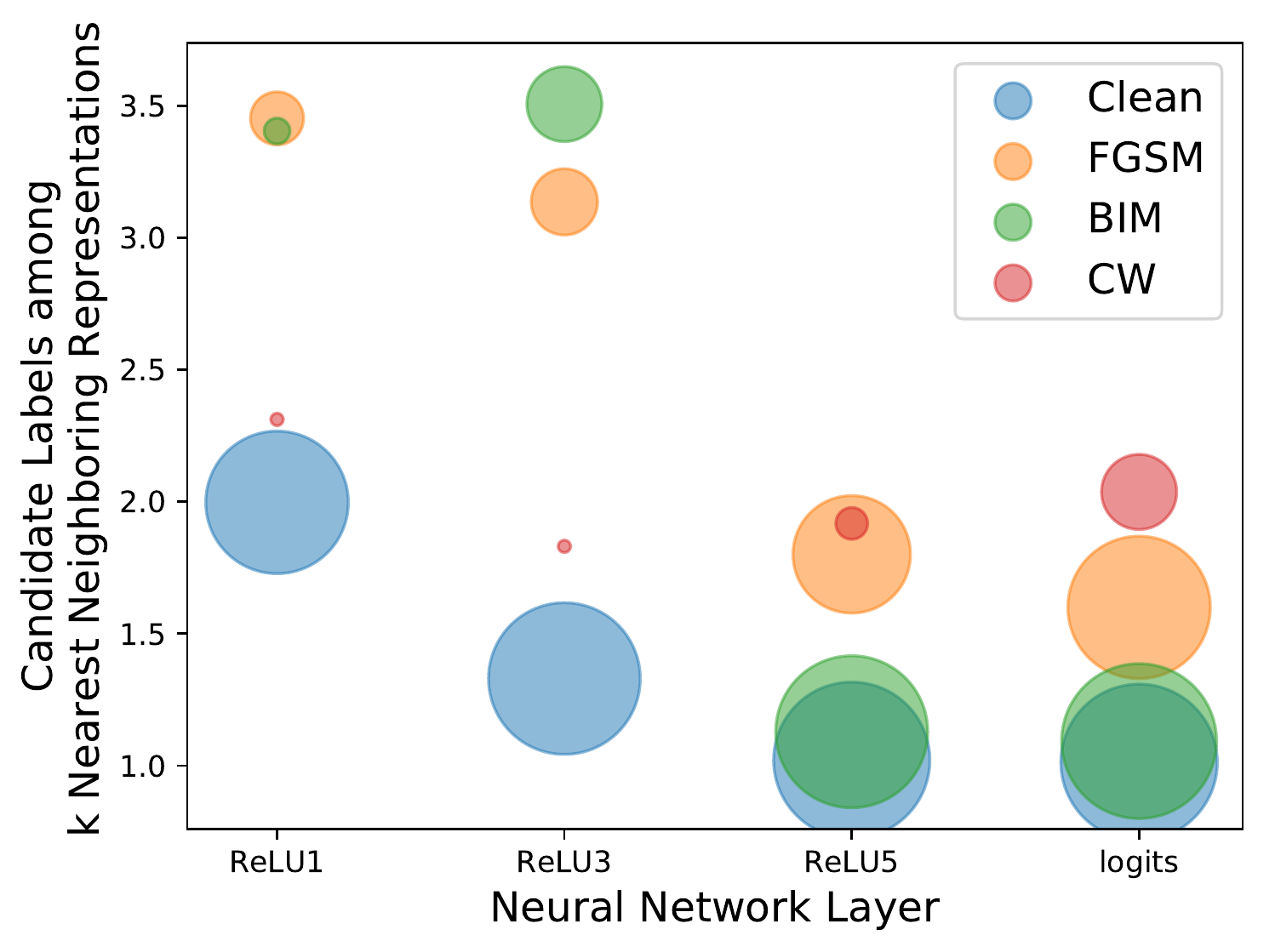}
	\caption{\textbf{Number of Candidate Labels among the $k$ Nearest Neighboring Representations}---Shown for MNIST. See Figure~\ref{fig:num-can-labels}
	for a detailed interpretation.}
\end{figure}

\begin{figure}[h] 
	\centering
	\includegraphics[width=\columnwidth]{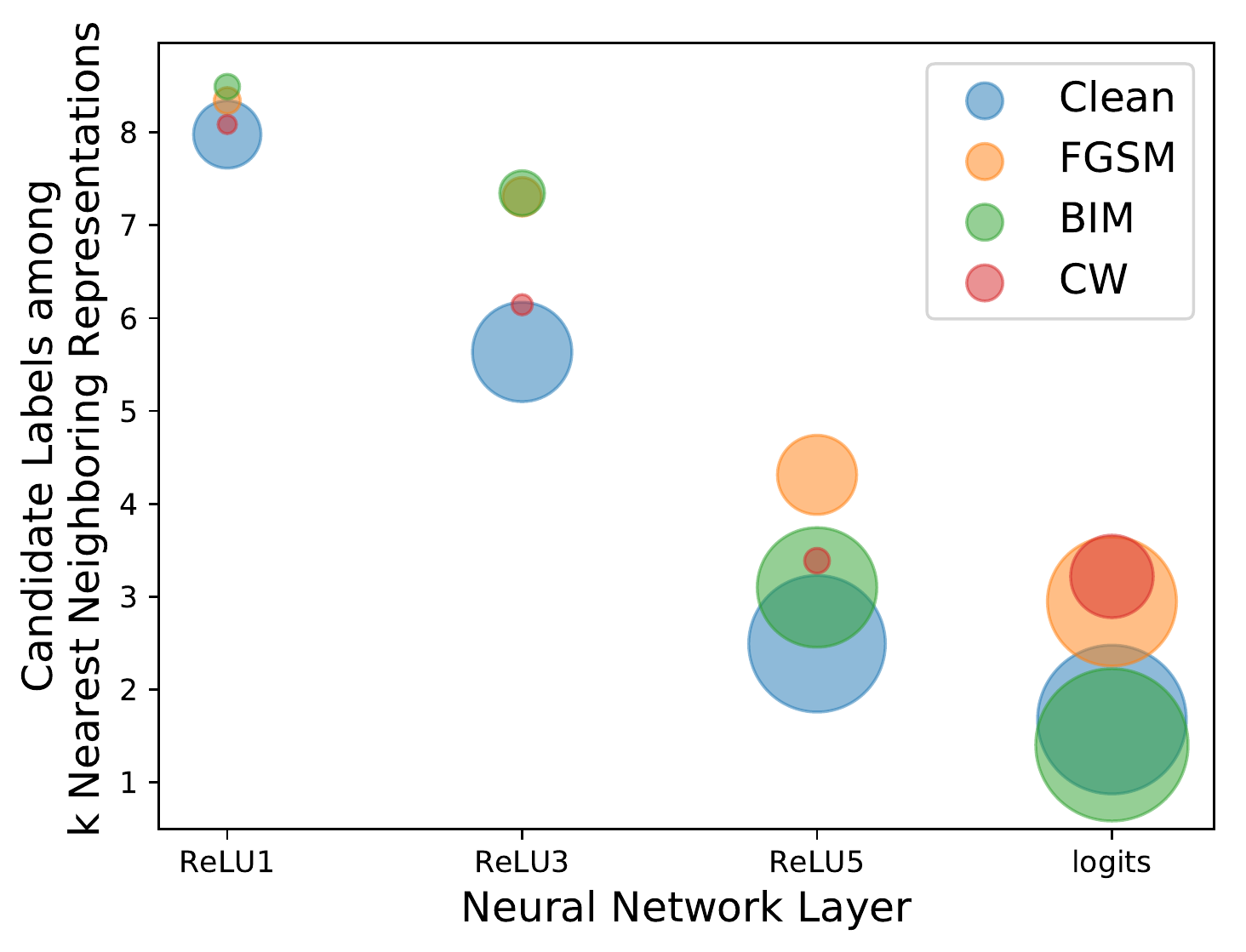}
	\caption{\textbf{Number of Candidate Labels among the $k$ Nearest Neighboring Representations}---Shown for SVHN. See Figure~\ref{fig:num-can-labels}
	for a detailed interpretation.}
\end{figure}

\newpage
\subsection{Feature Adversaries}

\begin{figure}[h] 
	\centering
	\includegraphics[width=0.9\columnwidth]{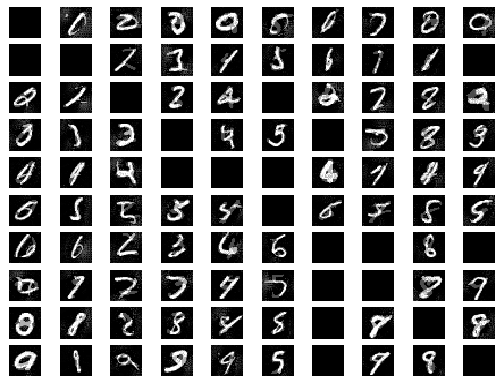}
	\caption{\textbf{Feature Adversarial Examples against our DkNN algorithm}---Shown for MNIST (see Figure~\ref{fig:feature-adv-ex} for details)}
\end{figure}

\end{document}